\documentclass{article}
\PassOptionsToPackage{numbers,compress}{natbib}
\usepackage[preprint]{neurips_2026}

\usepackage[utf8]{inputenc}
\usepackage[T1]{fontenc}
\usepackage{hyperref}
\usepackage{url}
\usepackage{booktabs}
\usepackage{amsfonts}
\usepackage{amsmath}
\usepackage{amssymb}
\usepackage{nicefrac}
\usepackage{microtype}
\usepackage[table]{xcolor}
\definecolor{bettergreen}{RGB}{0,128,0}
\usepackage{makecell}
\usepackage{graphicx}
\usepackage{subcaption}
\usepackage{multirow}
\usepackage{wrapfig}
\usepackage{enumitem}
\usepackage[capitalise,noabbrev]{cleveref}

\newcommand{\ie}{i.e.}
\newcommand{\eg}{e.g.}

\newcommand{\clapper}{\raisebox{-0.18em}{\includegraphics[height=1em]{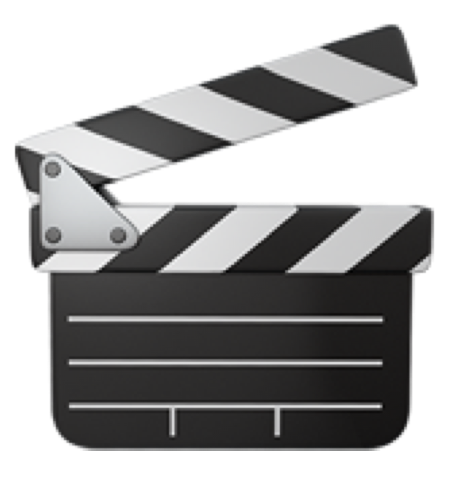}}}

\title{\clapper\;SocialDirector: Training-Free Social Interaction Control for Multi-Person Video Generation}

\author{%
  Liangyang Ouyang$^{1}$ \quad
  Ruicong Liu$^{1,2}$ \quad
  Caixin Kang$^{1}$ \quad
  Yifei Huang$^{1,2}$ \quad
  Yoichi Sato$^{1}$ \\[2pt]
  $^{1}$The University of Tokyo \quad
  $^{2}$Shanda AI Research Tokyo \\[2pt]
  \texttt{\{oyly, cxkang, ysato\}@iis.u-tokyo.ac.jp} \quad
  \texttt{\{ruicong.liu, yifei.huang\}@shanda.com}
}

\begin{document}

\maketitle

\begin{figure}[h]
    \centering
    \includegraphics[width=\linewidth]{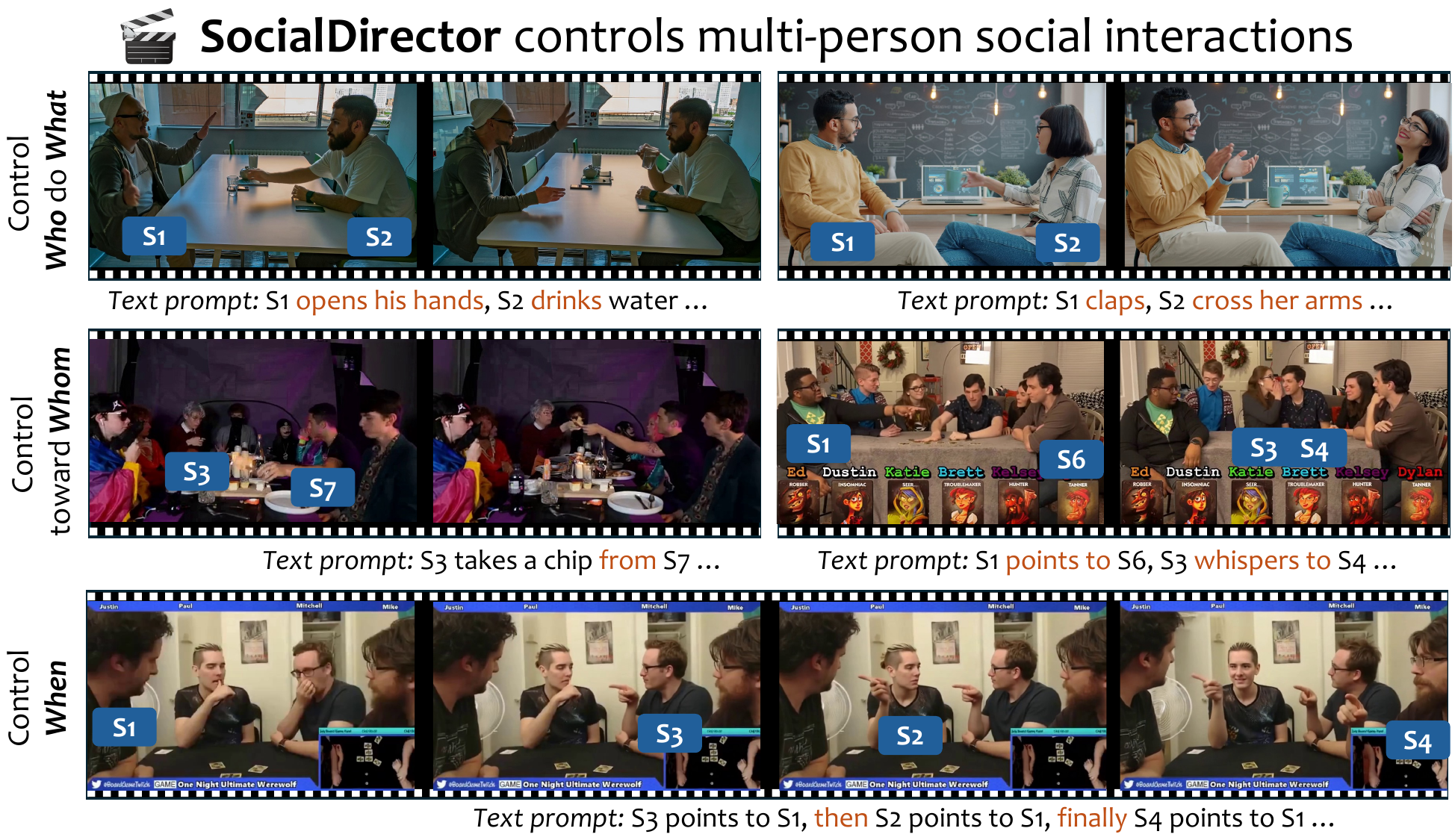}
    \vspace{-5mm}
    \caption{We propose \textbf{SocialDirector}, a training-free controller that enhances multi-person video generation with explicit control over social interactions. Based on a pretrained image-to-video diffusion transformer, SocialDirector controls \emph{who} performs \emph{what} action, \emph{when} each action occurs, and \emph{toward whom} it is directed, producing faithful interactions while preserving video quality.}
    \label{fig:teaser}
\end{figure}

\begin{abstract}
Video generation has advanced rapidly, producing photorealistic videos from text or image prompts. Meanwhile, film production and social robotics increasingly demand \emph{multi-person} videos with rich social interactions, including conversations, gestures, and coordinated actions. However, existing models offer no explicit control over interactions, such as \emph{who} performs \emph{which} action, \emph{when} it occurs, and \emph{toward whom} it is directed.
This often results in wrong person performing unintended actions (actor-action mismatch), disordered social dynamics, and wrong action targets.
To address these challenges, we present \textbf{SocialDirector}, a training-free interaction controller that enhances the generation model by modulating cross-attention maps. SocialDirector contains two modules: \emph{Social Actor Masking} and \emph{Directional Reweighting}. Social Actor Masking constrains each person's visual tokens to attend only to their own textual descriptions via a spatiotemporal mask, avoiding actor-action mismatch and disordered social dynamics.
Directional Reweighting amplifies attention to directional words (\eg, ``leftward'', ``right''), leading each action towards its intended target.
To evaluate generated social interactions, we annotate existing datasets with interaction descriptions and build a fully automated evaluation pipeline powered by open-source VLMs. Experiments on different video generation models show that SocialDirector significantly improves interaction fidelity and approaches the upper bound set by real videos.
\end{abstract}

\vspace{-5mm}
\section{Introduction}
\label{sec:intro}
\vspace{-2mm}

Diffusion-based video generation has advanced rapidly in recent years. Large-scale models such as Seedance \citep{gao2025seedance}, Wan~\citep{wan2025wan}, and LTX~\citep{hacohen2026ltx} can now synthesize photorealistic characters, motions, and camera movements from text or image prompts.
Building on this progress, studies have started extending to multi-person scenarios~\citep{multitalk,playmate2,huang2025bind,zhong2025anytalker,wang2025fantasyportrait}, motivated by applications in film production \citep{qiu2024moviecharacter,xu2025filmagent} and social robotics \citep{chen2025dancetogether,bharadhwaj2024gen2act}. Among these, social interactions, where individuals engage in conversations, gestures, and coordinated actions, have drawn growing attention.
Faithfully generating such interactions is essential for delivering immersive experiences.

However, existing generative models lack explicit control over interactions, such as \emph{who} performs \emph{what} action, \emph{when} it occurs, and \emph{toward whom} it is directed.
Consequently, generated videos often exhibit wrong person performing unintended actions (actor-action mismatch), disordered social dynamics, and wrong action targets.
We attribute this failure to \emph{attention leakage} in the cross-modal attention process: in a standard diffusion transformer (DiT) block, every visual token attends to all text tokens through cross-attention. As a result, the model imposes no spatial, temporal, or relational priors that tie a described action to its intended actor, moment, and target.

To address this limitation, we propose \textbf{SocialDirector}, an interaction controller that enhance the generation model by modulating its cross-attention maps. 
SocialDirector contains two training-free modules: \emph{Social Actor Masking} and \emph{Directional Reweighting}.
Social Actor Masking takes as input the first-frame image, per-person bounding boxes, and text prompt with timestamps.
From these, it constructs a spatiotemporal mask that restricts each person's visual tokens to attend only to their own textual description within the relevant time window.
This design eliminates attention leakage across person and time, avoiding actor-action mismatch and disordered social dynamics.
Directional Reweighting further amplifies attention to directional words (\eg, ``left'', ``right''), leading each action towards its intended target.
This module sharpens the model's sensitivity to directional cues, reducing mistargeted actions.
SocialDirector introduces no trainable parameters or architectural modifications, and can be seamlessly plugged into DiT-based video generation models.

Evaluating generated social interaction is an open challenge. Existing video quality metrics only measure visual quality, not whether actor, action, and target are correctly bound.
To this end, we annotate 149 samples from three existing multi-person datasets (MELD~\citep{poria2019meld}, MMSI~\citep{lee2024modeling}, and SocialGesture~\citep{cao2025socialgesture}) with structured interaction descriptions. 
Each description explicitly specifies the actor, action, target, timestamp, and per-person bounding boxes.
Based on these annotations, we build the first fully automated evaluation pipeline for social interactions.
This pipeline formulates the evaluation as binary video question answering, with answers aggregated by majority vote over three open-source VLMs.
Those answers are used to calculate three metrics: \emph{action accuracy}, \emph{target accuracy}, and \emph{stillness accuracy}.
These metrics measure, respectively, whether the actor and action match, whether it is directed at the correct target, and whether non-acting persons remain still.

We apply SocialDirector to two open-source generative models: Wan2.2~\citep{wan2025wan} and LTX2.3~\citep{hacohen2026ltx}.
On both backbones, our method consistently improves social interaction fidelity over the uncontrolled baselines, with gains of up to 3.9\%, 3.1\%, and 3.1\% on the three metrics. Notably, these results closely match the performance of a GT-oracle variant equipped with ground-truth annotations, suggesting that our generated interactions approach the upper bound set by real videos.
Our main contributions are summarized as follows:
\vspace{-1mm}

\begin{itemize}[leftmargin=1.5em]
    \item We propose SocialDirector, a training-free interaction controller that enhances the video diffusion model for multi-person video generation. It controls \emph{who} performs \emph{which} action, \emph{when} it occurs, and \emph{toward whom} it is directed through modulating cross-attention maps.
    \item We annotate existing datasets and build the first fully automated evaluation pipeline for social interaction video generation. It is powered by majority vote over open-source VLMs.
    \item Evaluations demonstrate that SocialDirector significantly improves social interaction fidelity and achieves state-of-the-art performance against competitive baselines.
\end{itemize}

\section{Related Works}
\label{sec:related}
\vspace{-2mm}

\subsection{Multimodal Social Interaction}
\label{sec:related_social}
\vspace{-1mm}

Human social interaction is inherently multimodal, spanning spoken language~\citep{jinnatural}, facial expressions~\citep{hyun2024smile}, gaze~\citep{zhou2024detecting,liu2021generalizing,liu2024pnp}, gestures~\citep{cao2025socialgesture,liu2024single}, and body movements~\citep{balazia2022bodily,liu2025sfhand}. The community has developed diverse tasks and benchmarks to study these phenomena, including video question answering~\citep{zadeh2019social,kong2025siv}, conversational modeling~\citep{ryan2023egocentric,lee2024modeling,jia2024audio}, speaker prediction~\citep{northcutt2020egocom,muller2021multimediate}, and behavior classification~\citep{cao2025socialgesture,ouyang2025leadership}, covering scenarios such as board games~\citep{lai2023werewolf,zhang2025multimind}, daily conversations~\citep{lei2018tvqa}, and meetings~\citep{kraaij2005ami}. Recent work has increasingly adopted multimodal foundation models for social interaction understanding~\citep{mathur2024advancing,lee2024towards,ouyang2025multi,kang2025can}. On the generation side, several methods synthesize multi-person 3D motion sequences conditioned on interaction categories or partner movements~\citep{yusocialgen,liang2024intergen,xu2024inter}. Compared with these understanding- and motion-level efforts, our work is the first to tackle \emph{controllable video generation} of multimodal social interactions.

\vspace{-1mm}
\subsection{Multi-Person Video Generation}
\label{sec:related_video}
\vspace{-1mm}

Human-centric video generation has been extensively studied, spanning talking video synthesis~\citep{wang2021one,chen2025midas}, portrait animation~\citep{guo2024liveportrait,chen2025echomimic}, and full-body motion generation~\citep{jiang2023text2performer,cheng2025wan,peng2025actavatar}. Extending to multi-person scenarios \citep{multitalk,facetoface,playmate2,huang2025bind,zhong2025anytalker,zhou2025evaltalker,zhou2026mtavg,wang2025fantasyportrait,chu2025unils,jinnatural,chen2025dancetogether} has recently attracted growing attention, where the central challenge has been correctly binding each person's audio to the corresponding individual in the scene. Existing approaches address this through label rotary position embedding~\citep{multitalk}, 3D-mask-based embedding routers~\citep{huang2025bind}, identity-aware audio-face cross attention~\citep{zhong2025anytalker}, and mask-guided classifier-free guidance~\citep{playmate2}. Unlike these audio-driven methods, our work focuses on the social interaction behaviors of multiple persons in generated videos---including action types, facial expressions, and especially interaction targets (\ie, who interacts with whom)---which have been largely overlooked by prior work. Furthermore, our approach is entirely training-free through cross-attention manipulation at inference time, whereas most existing multi-person video generation methods require dedicated multi-person training data and task-specific fine-tuning.

\vspace{-1mm}
\subsection{Cross-Attention Control in Diffusion Models}
\label{sec:related_attention}
\vspace{-1mm}

In diffusion models, cross-attention serves as the key mechanism for conditioning generation on external signals such as text prompts and reference images, applicable across both U-Net \citep{ronneberger2015u,rombach2022high} and Diffusion Transformer \citep{peebles2023scalable} backbones.
Cross-attention control has been widely adopted for text-to-image generation \citep{kim2023dense,chefer2023attend,rassin2023linguistic}, image editing \citep{hertzprompt,yang2023dynamic,alaluf2024cross}, layout-guided generation \citep{mao2023guided,chen2024training}, subject-driven synthesis \citep{wangms,wang2025ps,xiao2025fastcomposer}, noise optimization \citep{guo2024initno,ouyang2025lore}, and identity preservation \citep{zhang2025magicmirror,xie2025moca,yuan2025identity}.
Building on these works, SocialDirector applies cross-attention masking and reweighting to multi-person social interaction video generation, achieving per-person action binding, spatial identity grounding, and interaction target assignment.

\vspace{-1mm}
\section{Proposed Method}
\label{sec:method}
\vspace{-1mm}

\begin{figure*}[t]
\begin{center}
\includegraphics[width=\linewidth]{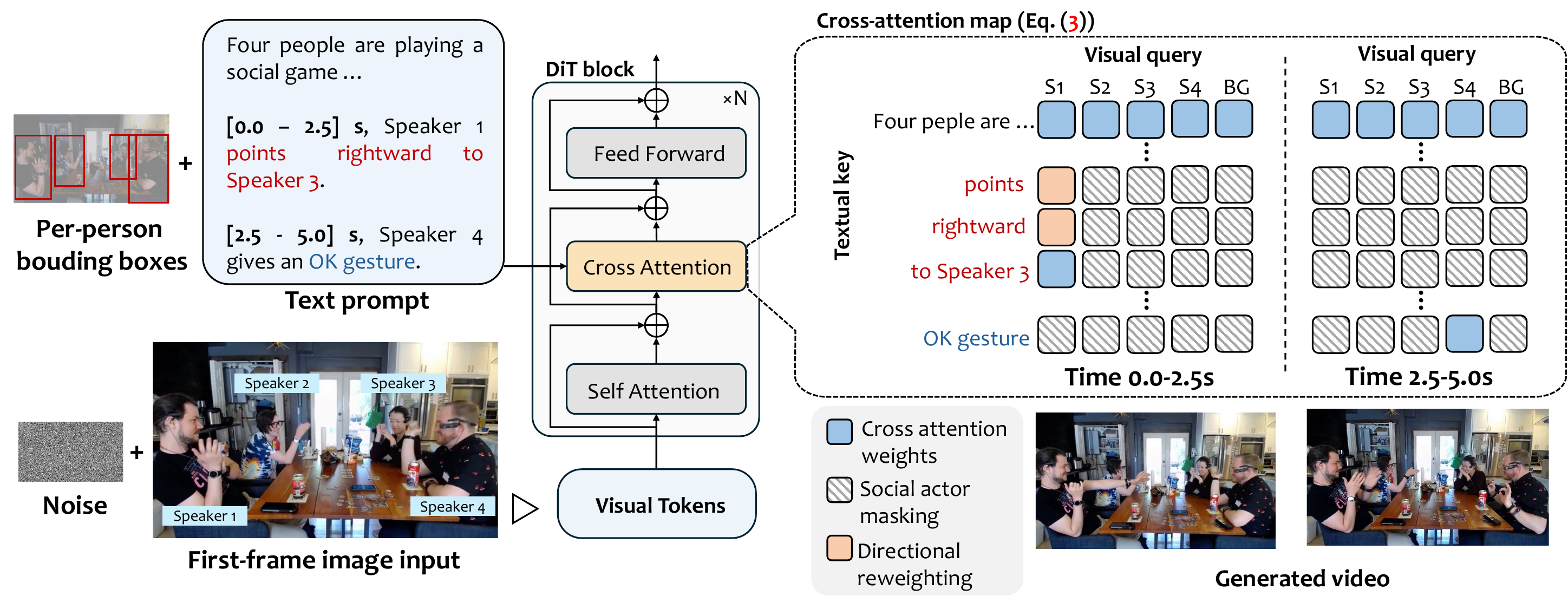}
\end{center}
\vspace{-4mm}
\caption{Overview of the proposed method. Given a first-frame image, per-person bounding boxes, and a structured text prompt describing per-person social events, our method generates multi-person videos with faithful social interactions. The generation is controlled through two modules modulating the cross-attention map: \emph{Social Actor Masking} and \emph{Directional Reweighting}.}
\vspace{-6mm}
\label{fig:method}
\end{figure*}

We propose \textbf{SocialDirector}, a training-free controller that enhances multi-person video generation with explicit control over social interactions. It consists of two modules that modulate the cross-attention map of each DiT block: \emph{Social Actor Masking} (\cref{sec:method_mask}), which applies a spatial-temporal mask that binds each social event to its own actor, and \emph{Directional Reweighting} (\cref{sec:method_reweight}), which guides actions toward correct targets. An overview is illustrated in \cref{fig:method}.

\subsection{Multimodal Social Interaction Video Generation}
\label{sec:method_task}

As shown in \cref{fig:method}, given an RGB first-frame image $I_0$ containing $N$ persons, a set of per-person bounding boxes $\mathcal{B}=\{b_s\mid s\in[1,N]\}$ of the image, and a text prompt $\mathcal{P}$, our social interaction video generation model outputs a video faithful to $\mathcal{P}$. Different from general multi-person generation, $\mathcal{P}$ is organized as a shared scene description followed by a sequence of $K$ \emph{structured social events}, each of the form ``$[t_1^k,t_2^k]$, person~$s$ performs action~$a$ (toward person~$s'$)''. The $[t_1^k,t_2^k]$ is the time window of the $k$-th event, $s\in[1,N]$ is the actor, $a$ is the action, and $s'$ is an optional interaction target. A faithful generation requires joint control over \emph{who} acts, \emph{what} action is performed, \emph{when} it occurs, and \emph{toward whom} it is directed.

\noindent\textbf{Attention leakage in standard DiTs.}
In a standard image-to-video DiT pipeline, $I_0$ and the added noise are tokenized into visual tokens $\mathcal{V}=\{v_i\in\mathbb{R}^d\mid i\in[1,T\times H\times W]\}$, and $\mathcal{P}$ is encoded into text tokens $\mathcal{L}=\{l_j\in\mathbb{R}^d\}$. Stacking these tokens as rows yields matrices $V$ and $L$, which serve respectively as queries and keys in the cross-attention of each DiT block: 

\vspace{-2mm}
\begin{equation}
\label{eq:attn}
\text{Attn}=\text{softmax}_l\!\left(\frac{VL^\top}{\sqrt{d}}\right).
\end{equation}
\vspace{-2mm}

This equation is agnostic to person identity and event time: a visual token belonging to one person can freely attend to text describing another person, at any moment in the video. As a result, actions are often performed by the wrong person, outside their annotated time window, or directed at the wrong target. We refer such failure as \emph{attention leakage}.
Attention leakage has been extensively studied in text-to-image generation, where cross-attention control is widely adopted for manipulating layout, attribute binding, and semantic composition in multi-subject outputs~\citep{yang2023dynamic,chefer2023attend,rassin2023linguistic}.

\vspace{-2mm}
\subsection{Social Actor Masking}
\label{sec:method_mask}
\vspace{-1mm}

To address attention leakage in multi-person video generation, we propose \emph{Social Actor Masking}, which binds each social event to its own actor in both space and time through a cross-attention mask. We encode $\mathcal{P}$ into a background token $\mathcal{L}_{\text{bg}}$ and per-event tokens $\mathcal{L}_{s,k}$. The background token $\mathcal{L}_{\text{bg}}$ describes the shared scene, while the per-event token $\mathcal{L}_{s,k}$ contains the action text of the $k$-th event associated with person~$s$. For visual tokens, we use the per-person bounding box $b_s$ and the event time window $[t_1^k,t_2^k]$ to localize the actor-specific spatiotemporal subset $\mathcal{V}_{s,k}\subset\mathcal{V}$ for each event. We then define a mask $M$ as:

\vspace{-3mm}
\begin{equation}
\label{eq:mask}
M_{v,l} =
\begin{cases}
0, & l \in \mathcal{L}_{\text{bg}}, \\
0, & l \in \mathcal{L}_{s,k}\ \text{and}\ v \in \mathcal{V}_{s,k}, \\
-\infty, & \text{otherwise.}
\end{cases}
\end{equation}
\vspace{-3mm}

As illustrated by the cross-attention map of \cref{fig:method}, the background span remains accessible to every visual query while each event span is gated to the visual tokens of its own actor within the relevant time window. The spatial dimension of $\mathcal{V}_{s,k}$ enforces \emph{who does what}, and its temporal dimension enforces \emph{when}, eliminating cross-person and cross-time attention leakage.

\vspace{-1mm}
\subsection{Directional Reweighting}
\label{sec:method_reweight}
\vspace{-1mm}



\cref{sec:method_mask} has no constraint on the action's target. As a result, mistargeted actions remain even when actor and time are correctly bound.
To address this, we introduce a \emph{Directional Reweighting} module.
Specifically, we first derive a direction (\eg, ``leftward'' or ``rightward'') from the actor's and target's bounding boxes, and insert the corresponding directional phrase (\eg, ``points rightward'' in \cref{fig:method}) into the text prompt.
We denote the set of all such directional tokens by $\mathcal{L}_{\text{dir}}$.
We then introduce a positive bias matrix $W$ that amplifies attention to these tokens, as illustrated by the cross-attention map of \cref{fig:method}. Together with $M$ in \cref{eq:mask}, $W$ is added to the cross-attention map:

\vspace{-3mm}
\begin{equation}
\label{eq:bias}
\begin{aligned}
W_{v,l} &=
\begin{cases}
\gamma\cdot\sqrt{d}, & l \in \mathcal{L}_{s,k}\cap\mathcal{L}_{\text{dir}}\ \text{and}\ v \in \mathcal{V}_{s,k}, \\
0, & \text{otherwise,}
\end{cases} \\[2pt]
\widehat{\text{Attn}} &= \text{softmax}_l\!\left(\frac{VL^\top}{\sqrt{d}} + M + W\right),
\end{aligned}
\end{equation}
\vspace{-3mm}

where the reweighting stength $\gamma$ is a pre-defined hyper-parameter. Scaling the bias by $\sqrt{d}$ matches the intrinsic scale of the unnormalized logits $VL^\top$, so that the $\gamma$ remains decoupled from the hidden dimension.
With $\gamma > 0$, this module sharpens the model's sensitivity to directional cues, guiding each action toward its intended target.

Both Social Actor Masking and Directional Reweighting operate purely on the cross-attention map at inference, introducing no additional parameters. SocialDirector is therefore plug-and-play for any DiT-based image-to-video backbone, requiring no architectural modification or fine-tuning.

\vspace{-1mm}
\section{Evaluation Pipeline}
\label{sec:eval}

We propose a fully automated evaluation pipeline for social interaction video generation.
Existing metrics are designed for video quality and identity consistency, failing to capture multi-person interaction fidelity.
Recent works \citep{zhou2025evaltalker,zhong2025anytalker} on multi-person video evaluation focus on audio-visual synchronization rather than social interactions, \eg, verifying the correctness of \emph{who} performs \emph{which} action toward \emph{whom}.
To fill this gap, we construct an evaluation dataset from existing social interaction benchmarks and employ open-source VLMs to assess generated videos.
An overview of the pipeline is provided in \cref{fig:eval}.

\subsection{Evaluation Dataset}
\label{sec:eval_dataset}

We collect ground-truth video clips from three publicly available multi-person interaction datasets, selecting clips with rich social interactions via the per-dataset procedures described below:

\textbf{MELD}~\citep{poria2019meld} is a multimodal multi-party dataset for emotion recognition in conversations, with annotations focused on emotional expressions during social exchanges. We select 19 five-second clips in which the scene and visible persons remain unchanged throughout.

\textbf{MMSI}~\citep{lee2024modeling} is a social interaction dataset with annotations of speaking behavior modeling including speaking targets. We select 50 clips that exhibit rich social dynamics (three or more speaking-target events within five seconds), retaining the original speaking-target and timestamp annotations.

\textbf{SocialGesture}~\citep{cao2025socialgesture} is a multi-person gesture recognition dataset captured from social game recordings. We randomly sample 80 annotated clips and ask annotators to re-annotate temporal timestamps within five seconds and additional social behaviors on top of the original gesture and target labels.

In total, the evaluation dataset contains 149 multi-person video clips, 674 annotated speakers, and 479 action events (299 with a specific interaction target), spanning 11 coarse action categories such as speaking, object interaction, and gestures. We additionally curate first-frame bounding-box annotations for every person, with manual corrections to ensure spatial accuracy. 

\vspace{-1mm}
\subsection{Social Interaction Metrics with VLMs}
\label{sec:eval_metrics}

A faithful social interaction video should correctly realize who performs what action, when it occurs, and toward whom it is directed. To evaluate these aspects, we design three complementary metrics targeting different dimensions of social interaction fidelity. \textbf{Action accuracy} verifies whether the designated person performs the correct action category within the annotated time window. For each annotated event, we generate a \emph{positive} query asking whether the correct person performs the expected action, and a \emph{negative} query asking whether another person performs that action, with expected answers ``yes'' and ``no'' respectively. \textbf{Target accuracy} further assesses whether an action is toward the correct target, through a similar positive/negative query pair that substitutes a wrong target. This metric is evaluated only on events with an explicit interaction target. \textbf{Stillness accuracy} measures whether persons with no annotated events correctly remain still, computed only on uninvolved speakers. Together, action accuracy evaluates ``who does what, when'', target accuracy evaluates ``toward whom'', and stillness accuracy penalizes actions that are not specified, providing a multi-dimensional social interaction evaluation. 

\begin{figure*}[t]
\centering
\includegraphics[width=\linewidth]{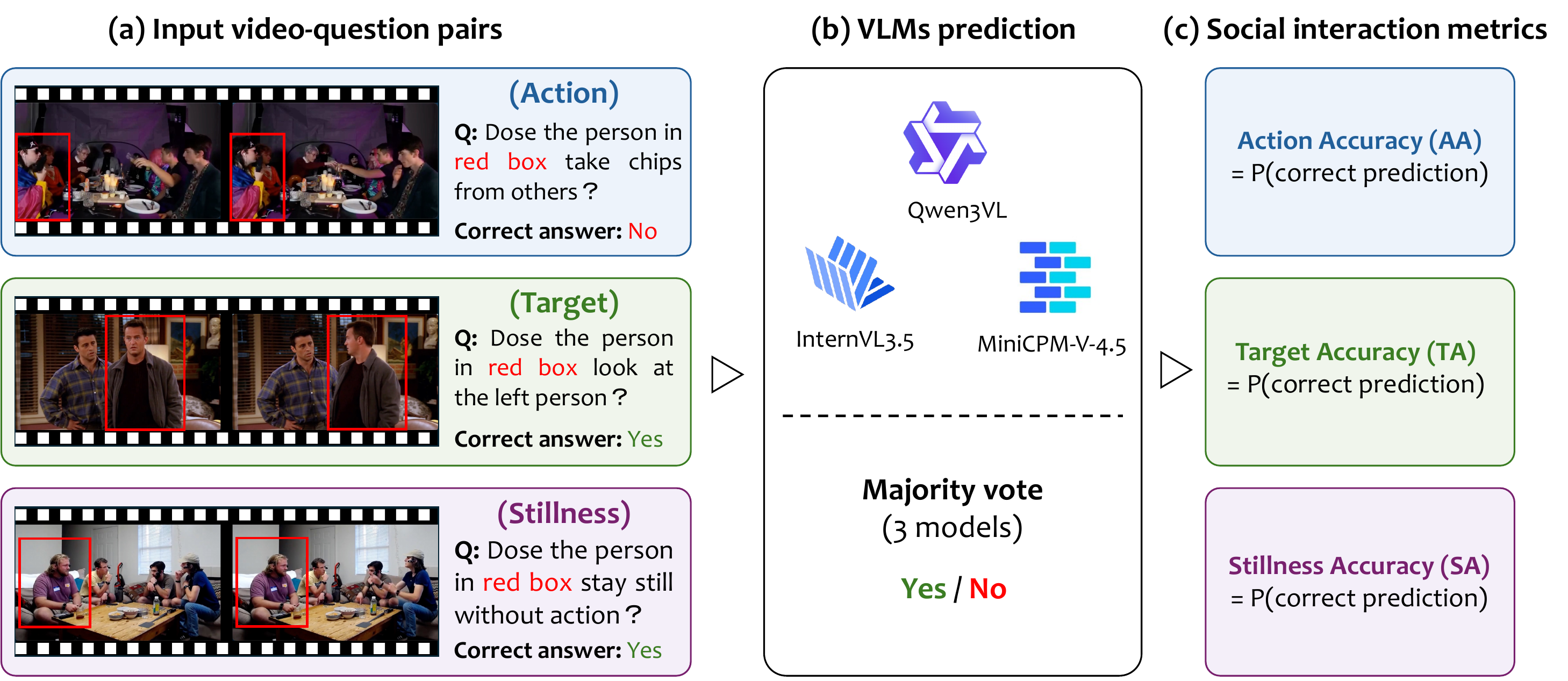}
\caption{Overview of our evaluation pipeline. (a)~Each annotated social event is converted into a video-question pair. (b)~Three VLMs independently answer each question (yes or no) and their predictions are aggregated via majority voting. (c)~The voting results, along with the correct answer, yield three social interaction metrics: Action Accuracy, Target Accuracy, and Stillness Accuracy.}
\label{fig:eval}
\end{figure*}

We formulate the above metrics binary visual question answering and evaluate each with three open-source VLMs. As illustrated in \cref{fig:eval}, each question is posed on a temporally trimmed sub-clip centered on the annotated event, so that the VLMs observe only the relevant time window. To specify the correct actor in multi-person scenes, we overlay a colored bounding box around the person of interest on the video \citep{li2025towards,yang2023fine,shtedritski2023does}. In total, the question set comprises 479 positive and 399 negative queries for action accuracy, 292 positive and 263 negative queries for target accuracy, and 314 queries for stillness accuracy. We employ three VLMs (Qwen3-VL~\citep{bai2025qwen3}, InternVL3.5~\citep{wang2025internvl3}, MiniCPM-V 4.5~\citep{yu2025minicpm}) and aggregate their answers via majority voting. More details about our evaluation metrics can be found in \cref{sec:appendix_eval}.

\vspace{-2mm}
\subsection{Video Quality Metrics}
\label{sec:eval_vq}

Beyond social interaction metrics, we adopt five complementary video quality metrics following previous multi-person video generation works~\citep{multitalk,zhong2025anytalker}: FVD~\citep{unterthiner2018towards} for distributional similarity to ground-truth videos, LPIPS~\citep{zhang2018unreasonable} for frame-level perceptual distance, ViCLIP~\citep{wanginternvid} similarity for text prompt alignment, DOVER~\citep{wu2023exploring} for aesthetic video quality, and DINOv2~\citep{oquabdinov2} cosine similarity for per-person identity preservation across frames~\citep{zhang2025magicmirror,yuan2025identity,xie2025moca}. 

\vspace{-2mm}
\section{Experiments}
\label{sec:experiments}

\subsection{Baseline Methods}
\label{sec:baselines}

We apply SocialDirector on top of two recent image-to-video DiTs: Wan2.2~\citep{wan2025wan} and LTX2.3~\citep{hacohen2026ltx}. We further compare against competitive multi-person video generation methods: MultiTalk~\citep{multitalk}, EchoMimic~v3~\citep{meng2026echomimicv3}, HunyuanVideoAvatar~\citep{chen2025hunyuanvideo}, Bind-Your-Avatar~\citep{huang2025bind}, AnyTalker~\citep{zhong2025anytalker}, and Playmate2~\citep{playmate2}. For fair comparison, all methods receive the same first-frame image and text prompt, where the text prompt explicitly describes the relative position and action of each speaker.
Methods that additionally accept per-person bounding-box inputs \citep{multitalk,playmate2,huang2025bind} are given identical boxes.
Since several of the baselines require an audio input, we supply a silent audio track to remove its impact. More details about baseline methods are provided in Appendix \cref{sec:appendix_baselines}.

\subsection{Implementation Details}
\label{sec:implementation}

Our main experiments use the pretrained Wan2.2-I2V-A14B~\citep{wan2025wan} and LTX-2.3-22B~\citep{hacohen2026ltx} checkpoints. We generate videos at $832\times 480$ resolution with 81 frames using both Wan2.2 and LTX-2.3. We use 40 denoising steps for Wan2.2 and 30 for LTX-2.3, following the officially recommended settings of each model. Our method is applied to all cross-attention layers of every DiT block. For Social Actor Masking, we empirically expand each person's bounding box by 15\% on each side to accommodate natural body movement beyond the first-frame annotation. For Directional Reweighting, we apply it to directional words and their verbs, \eg, ``pointing leftward'' and ``speaks to right''. The reweighting strength is set to $\gamma=0.5$. All methods are run three times with the same set of random seeds, and we report the averaged results. Each video is generated on a single NVIDIA L40S GPU.

\vspace{-2mm}
\subsection{Results}
\label{sec:results}

\begin{table*}[t]
\caption{Comparison of SocialDirector with baseline methods on our social interaction benchmark. \textcolor{bettergreen}{Green} indicates improvement over the base model; \textcolor{red}{red} indicates degradation.}

\label{tab:main}
\centering
\small
\resizebox{\linewidth}{!}{
\setlength{\tabcolsep}{1.2mm}
\begin{tabular}{lc|ccc|ccccc}
\Xhline{1.0pt}
\rowcolor[gray]{0.92} & 
 {\bf Box} & \multicolumn{3}{c|}{\bf Social Interaction Metrics} & \multicolumn{5}{c}{\bf Video Quality Metrics} \\
\rowcolor[gray]{0.92}
\multirow{-2}{*}{\bf Method} & {\bf Input} & Action Acc $\uparrow$ & Target Acc $\uparrow$ & Stillness Acc $\uparrow$ & FVD $\downarrow$ & LPIPS $\downarrow$ & ViCLIP $\uparrow$ & Vid Quality $\uparrow$ & ID Pres $\uparrow$ \\
\hline
\color{gray} GT Oracle & & \color{gray} 75.9 & \color{gray} 72.6 & \color{gray} 90.8 & \color{gray} --- & \color{gray} --- & \color{gray} 15.6 & \color{gray} 39.0 & \color{gray} 88.8 \\
\hline
EchoMimic v3 & & 60.1 & 65.5 & \textbf{98.0} & 5.26 & 17.6 & 14.4 & 35.9 & 90.8 \\
AnyTalker & & 64.1 & 66.3 & 92.9 & 5.40 & 16.8 & 15.2 & 36.3 & \textbf{91.7} \\
Bind-Your-Avatar & \checkmark & 63.6 & 69.5 & 72.2 & 20.1 & 31.2 & 16.4 & 28.4 & 66.9 \\
MultiTalk & \checkmark & 68.1 & 68.6 & 94.1 & 11.2 & 22.3 & 15.0 & 41.7 & 89.9 \\
HunyuanVideoAvatar & & 68.9 & 68.6 & 88.5 & 5.06 & 18.8 & 16.5 & 36.2 & 86.8 \\
\hline
LTX-2.3 & & 71.6 & 69.4 & 87.6 & 5.77 & 17.7 & 15.9 & 31.8 & 87.0 \\
\textbf{LTX-2.3 + SocialDirector} & \checkmark & 74.3\makebox[0pt][l]{\scriptsize\textcolor{bettergreen}{\,+2.7}} & 70.8\makebox[0pt][l]{\scriptsize\textcolor{bettergreen}{\,+1.4}} & 88.3\makebox[0pt][l]{\scriptsize\textcolor{bettergreen}{\,+0.7}} & 5.89\makebox[0pt][l]{\scriptsize\textcolor{red}{\,+0.12}} & 17.7\makebox[0pt][l]{\scriptsize\textcolor{bettergreen}{\,0.0}} & 16.4\makebox[0pt][l]{\scriptsize\textcolor{bettergreen}{\,+0.5}} & 32.4\makebox[0pt][l]{\scriptsize\textcolor{bettergreen}{\,+0.6}} & 86.7\makebox[0pt][l]{\scriptsize\textcolor{red}{\,$-$0.3}} \\
\Xhline{0.5pt}
Wan2.2 & & 72.2 & 69.2 & 85.1 & \textbf{4.21} & 17.2 & 16.1 & 43.3 & 84.1 \\
\textbf{Wan2.2 + SocialDirector} & \checkmark & \textbf{76.1}\makebox[0pt][l]{\scriptsize\textcolor{bettergreen}{\,+3.9}} & \textbf{72.3}\makebox[0pt][l]{\scriptsize\textcolor{bettergreen}{\,+3.1}} & 88.2\makebox[0pt][l]{\scriptsize\textcolor{bettergreen}{\,+3.1}} & 4.29\makebox[0pt][l]{\scriptsize\textcolor{red}{\,+0.08}} & \textbf{16.2}\makebox[0pt][l]{\scriptsize\textcolor{bettergreen}{\,$-$1.0}} & \textbf{16.8}\makebox[0pt][l]{\scriptsize\textcolor{bettergreen}{\,+0.7}} & \textbf{43.7}\makebox[0pt][l]{\scriptsize\textcolor{bettergreen}{\,+0.4}} & 85.1\makebox[0pt][l]{\scriptsize\textcolor{bettergreen}{\,+1.0}} \\
\Xhline{1.0pt}
\end{tabular}
}
\end{table*}

\paragraph{Comparison with baselines.}
As shown in \cref{tab:main}, applying SocialDirector to both Wan2.2 and LTX-2.3 yields consistent improvements on all three social interaction metrics. On Wan2.2, action accuracy improves by 3.9\%, target accuracy by 3.1\%, and stillness accuracy by 3.1\% over the base model, demonstrating the effectiveness of our method in resolving attention leakage and strengthening action-target binding. Compared with other multi-person video generation methods, SocialDirector achieves state-of-the-art performance on action and target accuracy. In particular, SocialDirector already matches the GT oracle on action and target accuracy on Wan2.2, reaching the upper bound set by real videos. Per-VLM (\cref{tab:appendix_per_vlm}) and per-seed (\cref{tab:appendix_per_seed}) analyses in the Appendix show that these gains hold across individual VLMs and random seeds, confirming that they are not artifacts of any particular random seed.
Meanwhile, the video quality metrics of SocialDirector are comparable to the base Wan2.2 and LTX-2.3, indicating that our method preserves visual quality.
 
For stillness accuracy and identity preservation, several audio-driven methods \citep{meng2026echomimicv3,zhong2025anytalker} exhibit abnormally high scores that even exceed the GT oracle. This occurs because these methods are trained primarily to align with audio streams and overlook text-driven controls. Their outputs tend to produce no action, which trivially inflates both metrics.
Other existing methods~\citep{multitalk,meng2026echomimicv3,chen2025hunyuanvideo,huang2025bind} perform poorly on social interaction metrics, as their implementations do not support scenes with three or more persons.

\begin{figure*}[t]
\centering
\includegraphics[width=\linewidth]{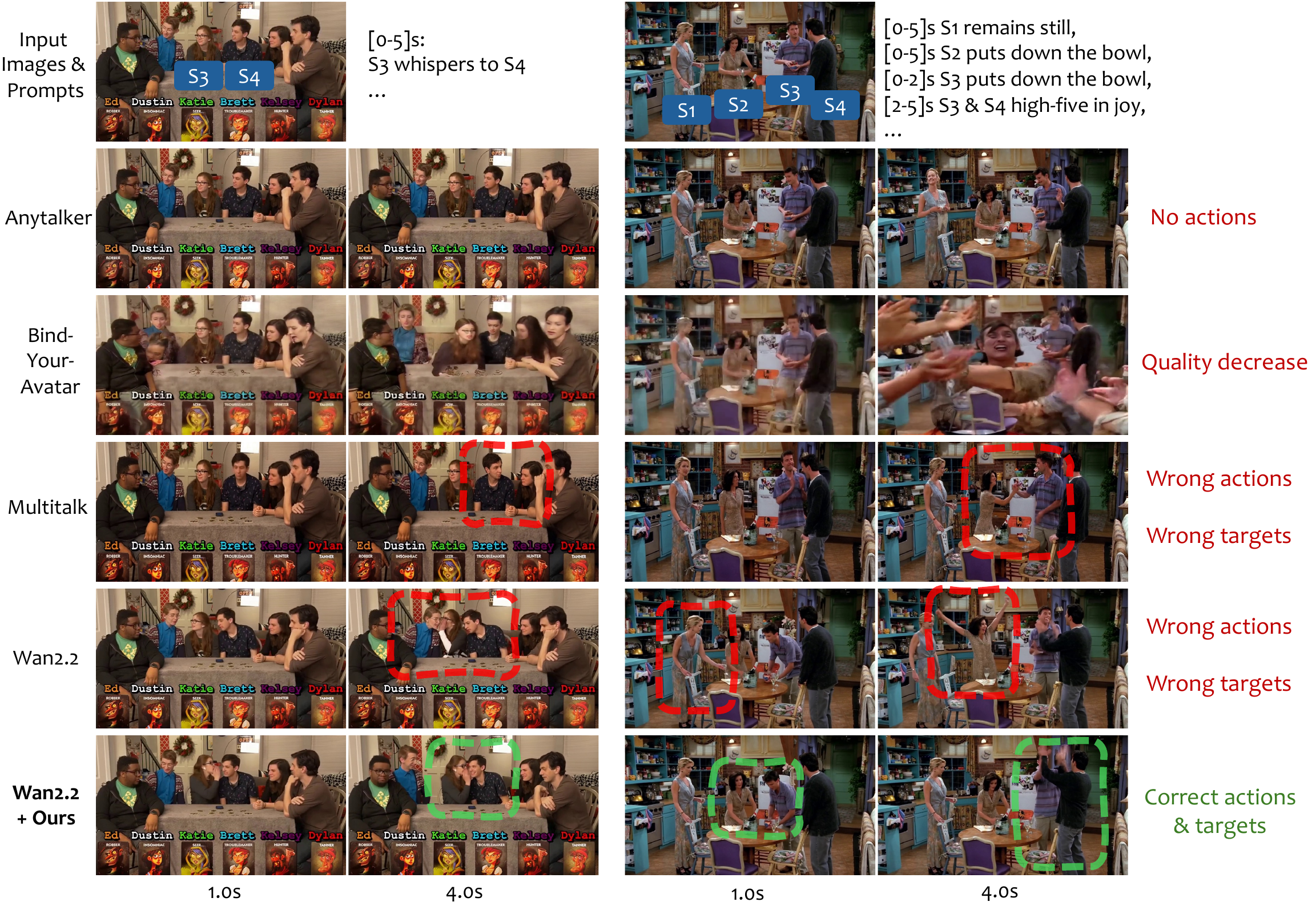}
\caption{Qualitative comparison with baseline methods on multi-person social interaction generation. All methods use the same random seed.}
\label{fig:vis}
\end{figure*}

\paragraph{Visualizations.}
Qualitative comparisons in \cref{fig:vis} highlight the advantages of SocialDirector over existing baselines. AnyTalker~\citep{zhong2025anytalker} remains largely insensitive to action controls from the text modality, resulting in nearly static videos. 
Bind-Your-Avatar~\citep{huang2025bind} suffers from significant degradation in visual quality when scaled from its two-person implementation to complex 4--6 person scenarios. Although MultiTalk~\citep{multitalk} and base Wan2.2 models generate the specified actions, they assign them to a wrong actor or direct them toward a wrong target, confirming the existence of attention leakage. In contrast, SocialDirector consistently generates faithful social interactions, precisely following the ``who,'' ``what,'' and ``to whom'' from text prompts. Qualitative comparisons against more methods are provided in Appendix \cref{sec:appendix_more_vis}.

\subsection{Ablations}
\label{sec:ablations}

\begin{table*}[t]
\centering
\small
\begin{minipage}[t]{0.48\linewidth}
\centering
\caption{Ablation on SocialDirector's components based on Wan2.2. $M$: Social Actor Masking, $W$: Directional Reweighting.}
\label{tab:ablation_components}
\resizebox{\linewidth}{!}{
\begin{tabular}{cc|ccc}
\Xhline{1.0pt}
\rowcolor[gray]{0.92}
$M$ & $W$ & Action Acc $\uparrow$ & Target Acc $\uparrow$ & Stillness Acc $\uparrow$ \\
\hline
 & & 72.2 & 69.2 & 85.1 \\
\checkmark & & 75.7 & 69.7 & 87.3 \\
 & \checkmark & 73.0 & 68.6 & 84.1 \\
\checkmark & \checkmark & \textbf{76.1} & \textbf{72.3} & \textbf{88.2} \\
\Xhline{1.0pt}
\end{tabular}
}
\end{minipage}
\hfill
\begin{minipage}[t]{0.48\linewidth}
\centering
\caption{Per-sample computational cost of SocialDirector.}
\label{tab:time}
\resizebox{\linewidth}{!}{
\begin{tabular}{l|cc}
\Xhline{1.0pt}
\rowcolor[gray]{0.92}
\bf Method & Time (s) & FLOPs ($\times10^{15}$) \\
\hline
Wan2.2 & 604.0 & 37.53 \\
Wan2.2 + Ours & 614.2\makebox[0pt][l]{\scriptsize\textcolor{red}{\,+1.7\%}} & 37.53 \\
LTX-2.3 & 180.4 & 33.79 \\
LTX-2.3 + Ours & 185.2\makebox[0pt][l]{\scriptsize\textcolor{red}{\,+2.7\%}} & 33.79 \\
\Xhline{1.0pt}
\end{tabular}
}
\end{minipage}
\end{table*}

\begin{figure*}[t]
\centering
\includegraphics[width=\linewidth]{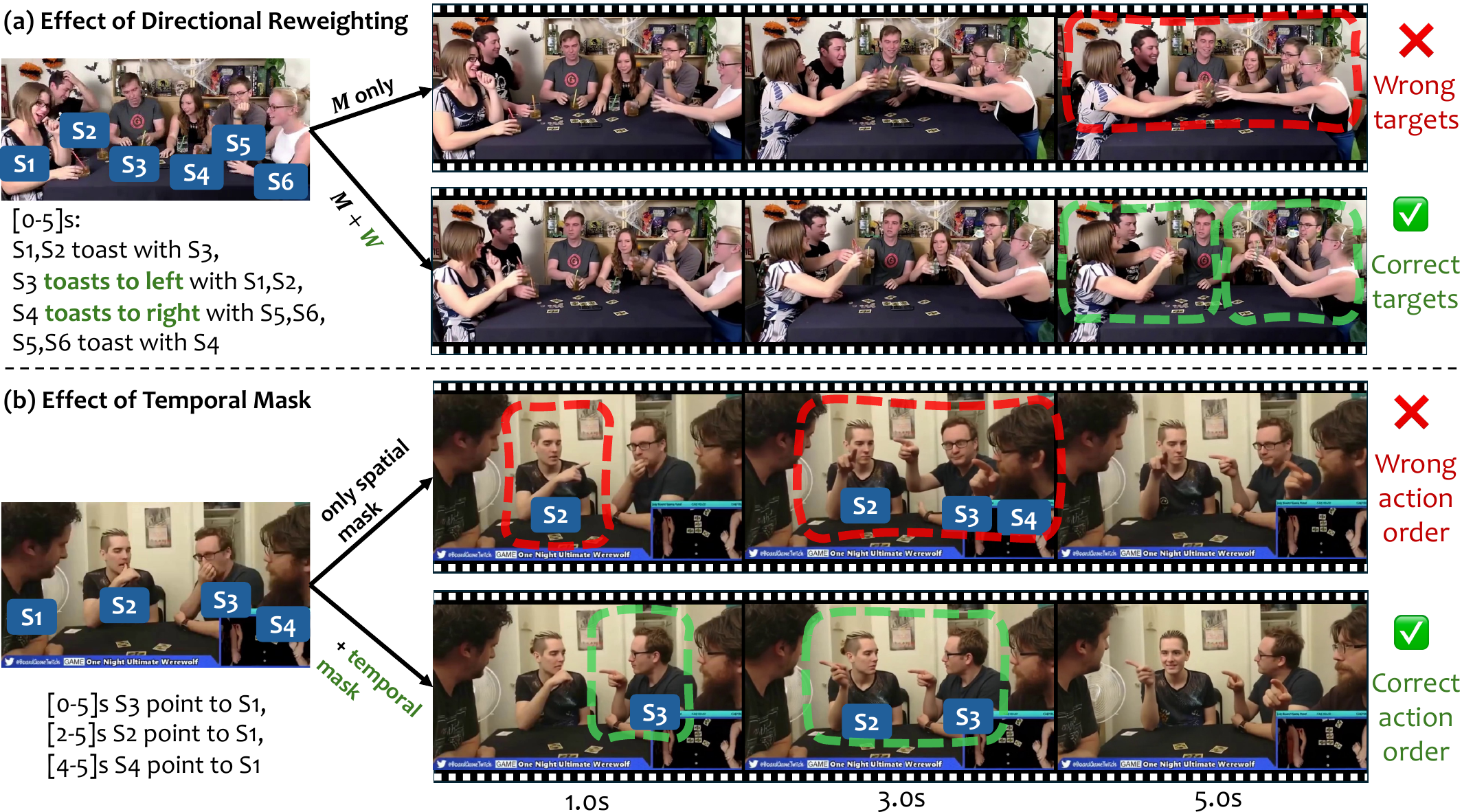}
\caption{Qualitative ablations of SocialDirector.}
\label{fig:ablation}
\end{figure*}

\paragraph{Effect of Attention Control Components.}We ablate Social Actor Masking and Directional Reweighting separately, \ie, $M$ and $W$ in \cref{eq:bias}. As shown in \cref{tab:ablation_components}, applying $M$ alone substantially improves action and stillness accuracy, confirming that attention leakage has been mitigated. However, gains in target accuracy remain limited. Adding $W$ further provides directional cues from the text side, boosting target accuracy by a large margin. 
In contrast, only applying $W$ offers no improvement, as it leaves attention leakage unresolved.

As visualized in \cref{fig:ablation} (a), $W$ guides actions toward intended targets. Specifically, when using only $M$, the model lacks awareness of target directions, resulting in an incorrect video where all speakers perform a collective toasting in the center. With $M+W$, Speaker 3 and Speaker 4 correctly toast to the left and right respectively, forming the two distinct interacting groups specified in the prompt. These two modules together achieve effective social interaction control.

\paragraph{Effect of Temporal Masking.}
We examine the effect of enabling the temporal dimension of Social Actor Masking. As shown in \cref{fig:ablation} (b), disabling temporal masking leads to a disorder in the sequence of events, where actions fail to respect their scheduled timings. Enabling temporal masking confines each action to its annotated time window and restores the correct event order, ensuring that the generated social dynamics faithfully follow the specified interaction timeline.

\paragraph{Computation Cost.}
\cref{tab:time} reports the inference overhead introduced by SocialDirector. On Wan2.2, the generation time increases by only 1.7\%, and on LTX-2.3 by 2.7\%. The additional FLOPs are negligible, as our method only performs element-wise additions on the cross-attention score matrices without introducing any new parameters or forward passes. This confirms the efficiency of our training-free approach.

\begin{table*}[t]
\centering
\small
\caption{Performance of different reweighting strength $\gamma$. All experiments are based on Wan2.2.}
\label{tab:ablation_gamma}
\resizebox{\linewidth}{!}{
\begin{tabular}{c|ccc|ccccc}
\Xhline{1.0pt}
\rowcolor[gray]{0.92}
 & \multicolumn{3}{c|}{\bf Social Interaction Metrics} & \multicolumn{5}{c}{\bf Video Quality Metrics} \\
\rowcolor[gray]{0.92}
\multirow{-2}{*}{$\gamma$} & Action Acc $\uparrow$ & Target Acc $\uparrow$ & Stillness Acc $\uparrow$ & FVD $\downarrow$ & LPIPS $\downarrow$ & ViCLIP $\uparrow$ & Vid Quality $\uparrow$ & ID Pres $\uparrow$ \\
\hline
0 & 75.7 & 69.7 & 87.3 & \textbf{4.11} & 16.4 & 16.4 & 43.6 & \textbf{85.8} \\
0.1 & 75.8 & 70.6 & 86.9 & 4.15 & 16.5 & 16.5 & 43.6 & 85.7 \\
0.3 & \textbf{76.2} & 71.6 & 87.6 & 4.15 & 16.3 & 16.1 & 43.6 & 85.6 \\
0.5 & 76.1 & \textbf{72.3} & 88.2 & 4.29 & \textbf{16.2} & \textbf{16.8} & \textbf{43.7} & 85.1 \\
0.7 & 75.6 & 69.4 & 88.5 & 4.84 & 17.2 & 16.2 & 41.7 & 83.2 \\
1.0 & 72.6 & 67.9 & \textbf{90.9} & 6.55 & 21.0 & 15.3 & 41.2 & 79.6 \\
\Xhline{1.0pt}
\end{tabular}
}
\vspace{-3mm}
\end{table*}

\paragraph{Reweighting Strength.}
\cref{tab:ablation_gamma} reports the sensitivity of our method to the reweighting strength $\gamma$. Smaller values provide insufficient directional bias, leading to lower target accuracy, whereas larger values over-amplify the directional tokens and degrade video quality. We select $\gamma=0.5$, which achieves the best overall performance.

\vspace{-2mm}
\section{Conclusion}
\label{sec:conclusion}
\vspace{-2mm}

This paper presents \textbf{SocialDirector}, a training-free controller that enhances pretrained image-to-video DiTs to faithfully generate multi-person social interaction videos. The proposed Social Actor Masking strengthens the binding between visual and textual tokens of the same actor across space and time. The proposed Directional Reweighting amplifies the attention to directional tokens, guiding each action toward its intended target. We further contribute a benchmark and the first fully automated VLM-based evaluation pipeline. Extensive experiments demonstrate that SocialDirector substantially improves interaction fidelity without compromising visual quality.

\noindent\textbf{Limitations and Future Directions.}
Three limitations of our current work point to natural directions for future research. First, SocialDirector operates in the visual-text domain and does not yet integrate the audio modality, falling short on speech-grounded behaviors such as lip synchronization. Second, our method is evaluated on short clips and does not support long video generation with sustained, multi-turn social dynamics. Third, our control relies on ground-truth bounding boxes and fails under heavy occlusion and large actor movements. Building on these limitations, future research includes incorporating audio for verbal and non-verbal signals, coupling with streaming avatar models for long-form generation, and replacing static box-based attention bias with dynamic positional-embedding-based conditioning.
Beyond this, a natural next step is to replace manual interaction scripts with an LLM-based social planner that automatically decides who should interact with whom and when. This moves video synthesis closer to autonomous, socially intelligent agents.

\bibliographystyle{plainnat}
\bibliography{references}

@article{wan2025wan,
  title={Wan: Open and advanced large-scale video generative models},
  author={Wan, Team and Wang, Ang and Ai, Baole and Wen, Bin and Mao, Chaojie and Xie, Chen-Wei and Chen, Di and Yu, Feiwu and Zhao, Haiming and Yang, Jianxiao and others},
  journal={arXiv preprint arXiv:2503.20314},
  year={2025}
}

@article{hacohen2026ltx,
  title={LTX-2: Efficient Joint Audio-Visual Foundation Model},
  author={HaCohen, Yoav and Brazowski, Benny and Chiprut, Nisan and Bitterman, Yaki and Kvochko, Andrew and Berkowitz, Avishai and Shalem, Daniel and Lifschitz, Daphna and Moshe, Dudu and Porat, Eitan and others},
  journal={arXiv preprint arXiv:2601.03233},
  year={2026}
}

@inproceedings{multitalk,
  title={Let Them Talk: Audio-Driven Multi-Person Conversational Video Generation},
  author={Kong, Zhe and Gao, Feng and Zhang, Yong and Kang, Zhuoliang and Wei, Xiaoming and Cai, Xunliang and Chen, Guanying and Luo, Wenhan},
  booktitle={The Thirty-ninth Annual Conference on Neural Information Processing Systems},
  year={2025}
}

@article{gao2025seedance,
  title={Seedance 1.0: Exploring the boundaries of video generation models},
  author={Gao, Yu and Guo, Haoyuan and Hoang, Tuyen and Huang, Weilin and Jiang, Lu and Kong, Fangyuan and Li, Huixia and Li, Jiashi and Li, Liang and Li, Xiaojie and others},
  journal={arXiv preprint arXiv:2506.09113},
  year={2025}
}

@article{facetoface,
  title={Face-to-Face: A Video Dataset for Multi-Person Interaction Modeling},
  author={Chu, Ernie and Patel, Vishal M},
  journal={arXiv preprint arXiv:2603.14794},
  year={2026}
}

@inproceedings{liu2021generalizing,
  title={Generalizing gaze estimation with outlier-guided collaborative adaptation},
  author={Liu, Yunfei and Liu, Ruicong and Wang, Haofei and Lu, Feng},
  booktitle={Proceedings of the IEEE/CVF international conference on computer vision},
  pages={3835--3844},
  year={2021}
}

@article{liu2024pnp,
  title={Pnp-ga+: Plug-and-play domain adaptation for gaze estimation using model variants},
  author={Liu, Ruicong and Liu, Yunfei and Wang, Haofei and Lu, Feng},
  journal={IEEE Transactions on Pattern Analysis and Machine Intelligence},
  volume={46},
  number={5},
  pages={3707--3721},
  year={2024},
  publisher={IEEE}
}

@inproceedings{liu2024single,
  title={Single-to-dual-view adaptation for egocentric 3d hand pose estimation},
  author={Liu, Ruicong and Ohkawa, Takehiko and Zhang, Mingfang and Sato, Yoichi},
  booktitle={Proceedings of the IEEE/CVF Conference on Computer Vision and Pattern Recognition},
  pages={677--686},
  year={2024}
}

@article{liu2025sfhand,
  title={SFHand: A Streaming Framework for Language-guided 3D Hand Forecasting and Embodied Manipulation},
  author={Liu, Ruicong and Huang, Yifei and Ouyang, Liangyang and Kang, Caixin and Sato, Yoichi},
  journal={arXiv preprint arXiv:2511.18127},
  year={2025}
}

@inproceedings{playmate2,
  title={Training-Free Multi-Character Audio-Driven Animation via Diffusion Transformer with Reward Feedback},
  author={Ma, Xingpei and Huang, Shenneng and Cai, Jiaran and Guan, Yuansheng and Zheng, Shen and Zhao, Hanfeng and Zhang, Qiang and Zhang, Shunsi},
  booktitle={Proceedings of the AAAI Conference on Artificial Intelligence},
  volume={40},
  number={10},
  pages={7818--7826},
  year={2026}
}

@article{zhou2025evaltalker,
  title={EvalTalker: Learning to Evaluate Real-Portrait-Driven Multi-Subject Talking Humans},
  author={Zhou, Yingjie and Zhu, Xilei and Ren, Siyu and Zhao, Ziyi and Wang, Ziwen and Wen, Farong and Zhou, Yu and Cao, Jiezhang and Min, Xiongkuo and Chen, Fengjiao and others},
  journal={arXiv preprint arXiv:2512.01340},
  year={2025}
}

@article{zhong2025anytalker,
  title={Anytalker: Scaling multi-person talking video generation with interactivity refinement},
  author={Zhong, Zhizhou and Ji, Yicheng and Kong, Zhe and Liu, Yiying and Wang, Jiarui and Feng, Jiasun and Liu, Lupeng and Wang, Xiangyi and Li, Yanjia and She, Yuqing and others},
  journal={arXiv preprint arXiv:2511.23475},
  year={2025}
}

@article{huang2025bind,
  title={Bind-your-avatar: Multi-talking-character video generation with dynamic 3d-mask-based embedding router},
  author={Huang, Yubo and Wang, Weiqiang and Zhao, Sirui and Xu, Tong and Liu, Lin and Chen, Enhong},
  journal={arXiv preprint arXiv:2506.19833},
  year={2025}
}

@article{zhou2026mtavg,
  title={MTAVG-Bench: A Comprehensive Benchmark for Evaluating Multi-Talker Dialogue-Centric Audio-Video Generation},
  author={Zhou, Yang-Hao and Li, Haitian and Lin, Rexar and Huang, Heyan and Zhou, Jinxing and Yuan, Changsen and Lan, Tian and Zhou, Ziqin and Li, Yudong and Xu, Jiajun and others},
  journal={arXiv preprint arXiv:2602.00607},
  year={2026}
}

@article{wang2025fantasyportrait,
  title={Fantasyportrait: Enhancing multi-character portrait animation with expression-augmented diffusion transformers},
  author={Wang, Qiang and Wang, Mengchao and Jiang, Fan and Fan, Yaqi and Qi, Yonggang and Xu, Mu},
  journal={arXiv preprint arXiv:2507.12956},
  year={2025}
}

@inproceedings{jinnatural,
  title={From Natural Alignment to Conditional Controllability in Multimodal Dialogue},
  author={Jin, Zeyu and Zhou, Songtao and Wang, Haoyu and Tian, Minghao and Yun, Kaifeng and Chen, Zhuo and Qin, Xiaoyu and Jia, Jia},
  booktitle={The Fourteenth International Conference on Learning Representations},
  year={2026}
}

@inproceedings{lee2024modeling,
  title={Modeling multimodal social interactions: new challenges and baselines with densely aligned representations},
  author={Lee, Sangmin and Lai, Bolin and Ryan, Fiona and Boote, Bikram and Rehg, James M},
  booktitle={Proceedings of the IEEE/CVF Conference on Computer Vision and Pattern Recognition},
  pages={14585--14595},
  year={2024}
}

@inproceedings{cao2025socialgesture,
  title={Socialgesture: Delving into multi-person gesture understanding},
  author={Cao, Xu and Virupaksha, Pranav and Jia, Wenqi and Lai, Bolin and Ryan, Fiona and Lee, Sangmin and Rehg, James M},
  booktitle={Proceedings of the Computer Vision and Pattern Recognition Conference},
  pages={19509--19519},
  year={2025}
}

@inproceedings{peebles2023scalable,
  title={Scalable diffusion models with transformers},
  author={Peebles, William and Xie, Saining},
  booktitle={Proceedings of the IEEE/CVF international conference on computer vision},
  pages={4195--4205},
  year={2023}
}

@inproceedings{hertzprompt,
  title={Prompt-to-Prompt Image Editing with Cross-Attention Control},
  author={Hertz, Amir and Mokady, Ron and Tenenbaum, Jay and Aberman, Kfir and Pritch, Yael and Cohen-Or, Daniel},
  booktitle={The Eleventh International Conference on Learning Representations},
  year={2023}
}

@article{yang2023dynamic,
  title={Dynamic prompt learning: Addressing cross-attention leakage for text-based image editing},
  author={Yang, Fei and Yang, Shiqi and Butt, Muhammad Atif and van de Weijer, Joost and others},
  journal={Advances in Neural Information Processing Systems},
  volume={36},
  pages={26291--26303},
  year={2023}
}

@inproceedings{poria2019meld,
  title={Meld: A multimodal multi-party dataset for emotion recognition in conversations},
  author={Poria, Soujanya and Hazarika, Devamanyu and Majumder, Navonil and Naik, Gautam and Cambria, Erik and Mihalcea, Rada},
  booktitle={Proceedings of the 57th annual meeting of the association for computational linguistics},
  pages={527--536},
  year={2019}
}

@inproceedings{hyun2024smile,
  title={SMILE: Multimodal Dataset for Understanding Laughter in Video with Language Models},
  author={Hyun, Lee and Sung-Bin, Kim and Han, Seungju and Yu, Youngjae and Oh, Tae-Hyun},
  booktitle={Findings of the Association for Computational Linguistics: NAACL 2024},
  pages={1149--1167},
  year={2024}
}

@article{zhou2024detecting,
  title={Detecting non-verbal speech and gaze behaviours with multimodal data and computer vision to interpret effective collaborative learning interactions},
  author={Zhou, Qi and Suraworachet, Wannapon and Cukurova, Mutlu},
  journal={Education and information technologies},
  volume={29},
  number={1},
  pages={1071--1098},
  year={2024},
  publisher={Springer}
}

@inproceedings{balazia2022bodily,
  title={Bodily behaviors in social interaction: Novel annotations and state-of-the-art evaluation},
  author={Balazia, Michal and M{\"u}ller, Philipp and T{\'a}nczos, {\'A}kos Levente and Liechtenstein, August von and Br{\'e}mond, Fran{\c{c}}ois},
  booktitle={Proceedings of the 30th ACM International Conference on Multimedia},
  pages={70--79},
  year={2022}
}

@inproceedings{zadeh2019social,
  title={Social-iq: A question answering benchmark for artificial social intelligence},
  author={Zadeh, Amir and Chan, Michael and Liang, Paul Pu and Tong, Edmund and Morency, Louis-Philippe},
  booktitle={Proceedings of the IEEE/CVF Conference on Computer Vision and Pattern Recognition},
  pages={8807--8817},
  year={2019}
}

@inproceedings{ryan2023egocentric,
  title={Egocentric auditory attention localization in conversations},
  author={Ryan, Fiona and Jiang, Hao and Shukla, Abhinav and Rehg, James M and Ithapu, Vamsi Krishna},
  booktitle={Proceedings of the IEEE/CVF Conference on Computer Vision and Pattern Recognition},
  pages={14663--14674},
  year={2023}
}

@inproceedings{jia2024audio,
  title={The audio-visual conversational graph: From an egocentric-exocentric perspective},
  author={Jia, Wenqi and Liu, Miao and Jiang, Hao and Ananthabhotla, Ishwarya and Rehg, James M and Ithapu, Vamsi Krishna and Gao, Ruohan},
  booktitle={Proceedings of the IEEE/CVF Conference on Computer Vision and Pattern Recognition},
  pages={26396--26405},
  year={2024}
}

@inproceedings{muller2021multimediate,
  title={Multimediate: Multi-modal group behaviour analysis for artificial mediation},
  author={M{\"u}ller, Philipp and Dietz, Michael and Schiller, Dominik and Thomas, Dominike and Zhang, Guanhua and Gebhard, Patrick and Andr{\'e}, Elisabeth and Bulling, Andreas},
  booktitle={Proceedings of the 29th ACM International Conference on Multimedia},
  pages={4878--4882},
  year={2021}
}

@article{lee2024towards,
  title={Towards social ai: A survey on understanding social interactions},
  author={Lee, Sangmin and Li, Minzhi and Lai, Bolin and Jia, Wenqi and Ryan, Fiona and Cao, Xu and Kara, Ozgur and Boote, Bikram and Shi, Weiyan and Yang, Diyi and others},
  journal={arXiv preprint arXiv:2409.15316},
  year={2024}
}

@article{ouyang2025multi,
  title={Multi-speaker Attention Alignment for Multimodal Social Interaction},
  author={Ouyang, Liangyang and Huang, Yifei and Zhang, Mingfang and Kang, Caixin and Furuta, Ryosuke and Sato, Yoichi},
  journal={arXiv preprint arXiv:2511.17952},
  year={2025}
}

@inproceedings{mathur2024advancing,
  title={Advancing Social Intelligence in AI Agents: Technical Challenges and Open Questions},
  author={Mathur, Leena and Liang, Paul Pu and Morency, Louis-Philippe},
  booktitle={Proceedings of the 2024 Conference on Empirical Methods in Natural Language Processing},
  pages={20541--20560},
  year={2024}
}

@article{kang2025can,
  title={Can MLLMs Read the Room? A Multimodal Benchmark for Assessing Deception in Multi-Party Social Interactions},
  author={Kang, Caixin and Huang, Yifei and Ouyang, Liangyang and Zhang, Mingfang and Liu, Ruicong and Sato, Yoichi},
  journal={arXiv preprint arXiv:2511.16221},
  year={2025}
}

@article{northcutt2020egocom,
  title={Egocom: A multi-person multi-modal egocentric communications dataset},
  author={Northcutt, Curtis G and Zha, Shengxin and Lovegrove, Steven and Newcombe, Richard},
  journal={IEEE Transactions on Pattern Analysis and Machine Intelligence},
  volume={45},
  number={6},
  pages={6783--6793},
  year={2020},
  publisher={IEEE}
}

@inproceedings{zhang2025multimind,
  title={Multimind: Enhancing werewolf agents with multimodal reasoning and theory of mind},
  author={Zhang, Zheng and Xiao, Nuoqian and Chai, Qi and Ye, Deheng and Wang, Hao},
  booktitle={Proceedings of the 33rd ACM International Conference on Multimedia},
  pages={5824--5833},
  year={2025}
}

@inproceedings{lai2023werewolf,
  title={Werewolf Among Us: Multimodal Resources for Modeling Persuasion Behaviors in Social Deduction Games},
  author={Lai, Bolin and Zhang, Hongxin and Liu, Miao and Pariani, Aryan and Ryan, Fiona and Jia, Wenqi and Hayati, Shirley Anugrah and Rehg, James and Yang, Diyi},
  booktitle={Findings of the Association for Computational Linguistics: ACL 2023},
  pages={6570--6588},
  year={2023}
}

@article{ouyang2025leadership,
  title={Leadership Assessment in Pediatric Intensive Care Unit Team Training},
  author={Ouyang, Liangyang and Sakai, Yuki and Furuta, Ryosuke and Nozawa, Hisataka and Matsui, Hikoro and Sato, Yoichi},
  journal={arXiv preprint arXiv:2505.24389},
  year={2025}
}

@inproceedings{lei2018tvqa,
  title={Tvqa: Localized, compositional video question answering},
  author={Lei, Jie and Yu, Licheng and Bansal, Mohit and Berg, Tamara},
  booktitle={Proceedings of the 2018 conference on empirical methods in natural language processing},
  pages={1369--1379},
  year={2018}
}

@inproceedings{kraaij2005ami,
  title={The AMI meeting corpus},
  author={Kraaij, Wessel and Hain, Thomas and Lincoln, Mike and Post, Wilfried},
  booktitle={Proc. International Conference on Methods and Techniques in Behavioral Research},
  pages={1--4},
  year={2005}
}

@article{kong2025siv,
  title={SIV-Bench: A Video Benchmark for Social Interaction Understanding and Reasoning},
  author={Kong, Fanqi and Zu, Weiqin and Chen, Xinyu and Yang, Yaodong and Zhu, Song-Chun and Feng, Xue},
  journal={arXiv preprint arXiv:2506.05425},
  year={2025}
}

@inproceedings{yusocialgen,
  title={SocialGen: Modeling Multi-Human Social Interaction with Language Models},
  author={Yu, Heng and Zhang, Juze and Chen, Changan and Xiang, Tiange and Fang, Yusu and Niebles, Juan Carlos and Adeli, Ehsan},
  booktitle={Thirteenth International Conference on 3D Vision},
  year={2026}
}

@article{liang2024intergen,
  title={Intergen: Diffusion-based multi-human motion generation under complex interactions},
  author={Liang, Han and Zhang, Wenqian and Li, Wenxuan and Yu, Jingyi and Xu, Lan},
  journal={International Journal of Computer Vision},
  volume={132},
  number={9},
  pages={3463--3483},
  year={2024},
  publisher={Springer}
}

@inproceedings{xu2024inter,
  title={Inter-x: Towards versatile human-human interaction analysis},
  author={Xu, Liang and Lv, Xintao and Yan, Yichao and Jin, Xin and Wu, Shuwen and Xu, Congsheng and Liu, Yifan and Zhou, Yizhou and Rao, Fengyun and Sheng, Xingdong and others},
  booktitle={Proceedings of the IEEE/CVF conference on computer vision and pattern recognition},
  pages={22260--22271},
  year={2024}
}

@article{chen2025dancetogether,
  title={Dancetogether! identity-preserving multi-person interactive video generation},
  author={Chen, Junhao and Chen, Mingjin and Xu, Jianjin and Li, Xiang and Dong, Junting and Sun, Mingze and Jiang, Puhua and Li, Hongxiang and Yang, Yuhang and Zhao, Hao and others},
  journal={arXiv preprint arXiv:2505.18078},
  year={2025}
}

@inproceedings{wang2021one,
  title={One-shot free-view neural talking-head synthesis for video conferencing},
  author={Wang, Ting-Chun and Mallya, Arun and Liu, Ming-Yu},
  booktitle={Proceedings of the IEEE/CVF conference on computer vision and pattern recognition},
  pages={10039--10049},
  year={2021}
}

@article{guo2024liveportrait,
  title={Liveportrait: Efficient portrait animation with stitching and retargeting control},
  author={Guo, Jianzhu and Zhang, Dingyun and Liu, Xiaoqiang and Zhong, Zhizhou and Zhang, Yuan and Wan, Pengfei and Zhang, Di},
  journal={arXiv preprint arXiv:2407.03168},
  year={2024}
}

@inproceedings{chen2025echomimic,
  title={Echomimic: Lifelike audio-driven portrait animations through editable landmark conditions},
  author={Chen, Zhiyuan and Cao, Jiajiong and Chen, Zhiquan and Li, Yuming and Ma, Chenguang},
  booktitle={Proceedings of the AAAI Conference on Artificial Intelligence},
  volume={39},
  number={3},
  pages={2403--2410},
  year={2025}
}

@article{chen2025hunyuanvideo,
  title={Hunyuanvideo-avatar: High-fidelity audio-driven human animation for multiple characters},
  author={Chen, Yi and Liang, Sen and Zhou, Zixiang and Huang, Ziyao and Ma, Yifeng and Tang, Junshu and Lin, Qin and Zhou, Yuan and Lu, Qinglin},
  journal={arXiv preprint arXiv:2505.20156},
  year={2025}
}

@inproceedings{meng2026echomimicv3,
  title={Echomimicv3: 1.3 b parameters are all you need for unified multi-modal and multi-task human animation},
  author={Meng, Rang and Wang, Yan and Wu, Weipeng and Zheng, Ruobing and Li, Yuming and Ma, Chenguang},
  booktitle={Proceedings of the AAAI Conference on Artificial Intelligence},
  volume={40},
  number={10},
  pages={8008--8015},
  year={2026}
}

@inproceedings{jiang2023text2performer,
  title={Text2performer: Text-driven human video generation},
  author={Jiang, Yuming and Yang, Shuai and Koh, Tong Liang and Wu, Wayne and Loy, Chen Change and Liu, Ziwei},
  booktitle={Proceedings of the IEEE/CVF International Conference on Computer Vision},
  pages={22747--22757},
  year={2023}
}

@article{cheng2025wan,
  title={Wan-animate: Unified character animation and replacement with holistic replication},
  author={Cheng, Gang and Gao, Xin and Hu, Li and Hu, Siqi and Huang, Mingyang and Ji, Chaonan and Li, Ju and Meng, Dechao and Qi, Jinwei and Qiao, Penchong and others},
  journal={arXiv preprint arXiv:2509.14055},
  year={2025}
}

@article{chu2025unils,
  title={UniLS: End-to-End Audio-Driven Avatars for Unified Listening and Speaking},
  author={Chu, Xuangeng and Liu, Ruicong and Huang, Yifei and Liu, Yun and Peng, Yichen and Zheng, Bo},
  journal={arXiv preprint arXiv:2512.09327},
  year={2025}
}

@article{peng2025actavatar,
  title={ActAvatar: Temporally-Aware Precise Action Control for Talking Avatars},
  author={Peng, Ziqiao and Chen, Yi and Ma, Yifeng and Zhang, Guozhen and Sun, Zhiyao and Zhou, Zixiang and Zhang, Youliang and Zhou, Zhengguang and Fan, Zhaoxin and Liu, Hongyan and others},
  journal={arXiv preprint arXiv:2512.19546},
  year={2025}
}

@inproceedings{rombach2022high,
  title={High-resolution image synthesis with latent diffusion models},
  author={Rombach, Robin and Blattmann, Andreas and Lorenz, Dominik and Esser, Patrick and Ommer, Bj{\"o}rn},
  booktitle={Proceedings of the IEEE/CVF conference on computer vision and pattern recognition},
  pages={10684--10695},
  year={2022}
}

@inproceedings{ronneberger2015u,
  title={U-net: Convolutional networks for biomedical image segmentation},
  author={Ronneberger, Olaf and Fischer, Philipp and Brox, Thomas},
  booktitle={International Conference on Medical image computing and computer-assisted intervention},
  pages={234--241},
  year={2015},
  organization={Springer}
}

@article{ouyang2025lore,
  title={LORE: Latent Optimization for Precise Semantic Control in Rectified Flow-based Image Editing},
  author={Ouyang, Liangyang and Mao, Jiafeng},
  journal={arXiv preprint arXiv:2508.03144},
  year={2025}
}

@inproceedings{mao2023guided,
  title={Guided image synthesis via initial image editing in diffusion model},
  author={Mao, Jiafeng and Wang, Xueting and Aizawa, Kiyoharu},
  booktitle={Proceedings of the 31st ACM International Conference on Multimedia},
  pages={5321--5329},
  year={2023}
}

@inproceedings{zhang2025magicmirror,
  title={Magicmirror: Id-preserved video generation in video diffusion transformers},
  author={Zhang, Yuechen and Liu, Yaoyang and Xia, Bin and Peng, Bohao and Yan, Zexin and Lo, Eric and Jia, Jiaya},
  booktitle={Proceedings of the IEEE/CVF International Conference on Computer Vision},
  pages={14464--14474},
  year={2025}
}

@inproceedings{wangms,
  title={MS-Diffusion: Multi-subject Zero-shot Image Personalization with Layout Guidance},
  author={Wang, Xierui and Fu, Siming and Huang, Qihan and He, Wanggui and Jiang, Hao},
  booktitle={The Thirteenth International Conference on Learning Representations},
  year={2025}
}

@inproceedings{wang2025ps,
  title={PS-Diffusion: Photorealistic Subject-Driven Image Editing with Disentangled Control and Attention},
  author={Wang, Weicheng and Jia, Guoli and Zhang, Zhongqi and Lin, Liang and Yang, Jufeng},
  booktitle={2025 IEEE/CVF Conference on Computer Vision and Pattern Recognition (CVPR)},
  pages={18302--18312},
  year={2025},
  organization={IEEE Computer Society}
}

@inproceedings{kim2023dense,
  title={Dense text-to-image generation with attention modulation},
  author={Kim, Yunji and Lee, Jiyoung and Kim, Jin-Hwa and Ha, Jung-Woo and Zhu, Jun-Yan},
  booktitle={Proceedings of the IEEE/CVF International Conference on Computer Vision},
  pages={7701--7711},
  year={2023}
}

@inproceedings{alaluf2024cross,
  title={Cross-image attention for zero-shot appearance transfer},
  author={Alaluf, Yuval and Garibi, Daniel and Patashnik, Or and Averbuch-Elor, Hadar and Cohen-Or, Daniel},
  booktitle={ACM SIGGRAPH 2024 conference papers},
  pages={1--12},
  year={2024}
}

@article{chefer2023attend,
  title={Attend-and-excite: Attention-based semantic guidance for text-to-image diffusion models},
  author={Chefer, Hila and Alaluf, Yuval and Vinker, Yael and Wolf, Lior and Cohen-Or, Daniel},
  journal={ACM transactions on Graphics (TOG)},
  volume={42},
  number={4},
  pages={1--10},
  year={2023},
  publisher={ACM New York, NY, USA}
}

@inproceedings{chen2024training,
  title={Training-free layout control with cross-attention guidance},
  author={Chen, Minghao and Laina, Iro and Vedaldi, Andrea},
  booktitle={Proceedings of the IEEE/CVF winter conference on applications of computer vision},
  pages={5343--5353},
  year={2024}
}

@article{chen2025midas,
  title={Midas: Multimodal interactive digital-human synthesis via real-time autoregressive video generation},
  author={Chen, Ming and Cui, Liyuan and Zhang, Wenyuan and Zhang, Haoxian and Zhou, Yan and Li, Xiaohan and Tang, Songlin and Liu, Jiwen and Liao, Borui and Chen, Hejia and others},
  journal={arXiv preprint arXiv:2508.19320},
  year={2025}
}

@article{xiao2025fastcomposer,
  title={Fastcomposer: Tuning-free multi-subject image generation with localized attention},
  author={Xiao, Guangxuan and Yin, Tianwei and Freeman, William T and Durand, Fr{\'e}do and Han, Song},
  journal={International Journal of Computer Vision},
  volume={133},
  number={3},
  pages={1175--1194},
  year={2025},
  publisher={Springer}
}

@article{rassin2023linguistic,
  title={Linguistic binding in diffusion models: Enhancing attribute correspondence through attention map alignment},
  author={Rassin, Royi and Hirsch, Eran and Glickman, Daniel and Ravfogel, Shauli and Goldberg, Yoav and Chechik, Gal},
  journal={Advances in Neural Information Processing Systems},
  volume={36},
  pages={3536--3559},
  year={2023}
}

@inproceedings{guo2024initno,
  title={Initno: Boosting text-to-image diffusion models via initial noise optimization},
  author={Guo, Xiefan and Liu, Jinlin and Cui, Miaomiao and Li, Jiankai and Yang, Hongyu and Huang, Di},
  booktitle={Proceedings of the IEEE/CVF Conference on Computer Vision and Pattern Recognition},
  pages={9380--9389},
  year={2024}
}

@inproceedings{yuan2025identity,
  title={Identity-preserving text-to-video generation by frequency decomposition},
  author={Yuan, Shenghai and Huang, Jinfa and He, Xianyi and Ge, Yunyang and Shi, Yujun and Chen, Liuhan and Luo, Jiebo and Yuan, Li},
  booktitle={Proceedings of the Computer Vision and Pattern Recognition Conference},
  pages={12978--12988},
  year={2025}
}

@inproceedings{xie2025moca,
  title={Moca: Identity-preserving text-to-video generation via mixture of cross attention},
  author={Xie, Qi and Ma, Yongjia and Di, Donglin and Gao, Xuehao and Yang, Xun},
  booktitle={Proceedings of the 7th ACM International Conference on Multimedia in Asia},
  pages={1--8},
  year={2025}
}

@article{yu2025minicpm,
  title={Minicpm-v 4.5: Cooking efficient mllms via architecture, data, and training recipe},
  author={Yu, Tianyu and Wang, Zefan and Wang, Chongyi and Huang, Fuwei and Ma, Wenshuo and He, Zhihui and Cai, Tianchi and Chen, Weize and Huang, Yuxiang and Zhao, Yuanqian and others},
  journal={arXiv preprint arXiv:2509.18154},
  year={2025}
}

@article{bai2025qwen3,
  title={Qwen3-vl technical report},
  author={Bai, Shuai and Cai, Yuxuan and Chen, Ruizhe and Chen, Keqin and Chen, Xionghui and Cheng, Zesen and Deng, Lianghao and Ding, Wei and Gao, Chang and Ge, Chunjiang and others},
  journal={arXiv preprint arXiv:2511.21631},
  year={2025}
}

@article{wang2025internvl3,
  title={Internvl3. 5: Advancing open-source multimodal models in versatility, reasoning, and efficiency},
  author={Wang, Weiyun and Gao, Zhangwei and Gu, Lixin and Pu, Hengjun and Cui, Long and Wei, Xingguang and Liu, Zhaoyang and Jing, Linglin and Ye, Shenglong and Shao, Jie and others},
  journal={arXiv preprint arXiv:2508.18265},
  year={2025}
}

@article{unterthiner2018towards,
  title={Towards accurate generative models of video: A new metric \& challenges},
  author={Unterthiner, Thomas and Van Steenkiste, Sjoerd and Kurach, Karol and Marinier, Raphael and Michalski, Marcin and Gelly, Sylvain},
  journal={arXiv preprint arXiv:1812.01717},
  year={2018}
}

@inproceedings{zhang2018unreasonable,
  title={The unreasonable effectiveness of deep features as a perceptual metric},
  author={Zhang, Richard and Isola, Phillip and Efros, Alexei A and Shechtman, Eli and Wang, Oliver},
  booktitle={Proceedings of the IEEE conference on computer vision and pattern recognition},
  pages={586--595},
  year={2018}
}

@inproceedings{wu2023exploring,
  title={Exploring video quality assessment on user generated contents from aesthetic and technical perspectives},
  author={Wu, Haoning and Zhang, Erli and Liao, Liang and Chen, Chaofeng and Hou, Jingwen and Wang, Annan and Sun, Wenxiu and Yan, Qiong and Lin, Weisi},
  booktitle={Proceedings of the IEEE/CVF international conference on computer vision},
  pages={20144--20154},
  year={2023}
}

@article{oquabdinov2,
  title={DINOv2: Learning Robust Visual Features without Supervision},
  author={Oquab, Maxime and Darcet, Timoth{\'e}e and Moutakanni, Th{\'e}o and Vo, Huy V and Szafraniec, Marc and Khalidov, Vasil and Fernandez, Pierre and HAZIZA, Daniel and Massa, Francisco and El-Nouby, Alaaeldin and others},
  journal={Transactions on Machine Learning Research},
  year={2024}
}

@inproceedings{wanginternvid,
  title={InternVid: A Large-scale Video-Text Dataset for Multimodal Understanding and Generation},
  author={Wang, Yi and He, Yinan and Li, Yizhuo and Li, Kunchang and Yu, Jiashuo and Ma, Xin and Li, Xinhao and Chen, Guo and Chen, Xinyuan and Wang, Yaohui and others},
  booktitle={The Twelfth International Conference on Learning Representations},
  year={2024}
}

@article{li2025towards,
  title={Towards online multi-modal social interaction understanding},
  author={Li, Xinpeng and Deng, Shijian and Lai, Bolin and Pian, Weiguo and Rehg, James M and Tian, Yapeng},
  journal={arXiv preprint arXiv:2503.19851},
  year={2025}
}

@article{yang2023fine,
  title={Fine-grained visual prompting},
  author={Yang, Lingfeng and Wang, Yueze and Li, Xiang and Wang, Xinlong and Yang, Jian},
  journal={Advances in Neural Information Processing Systems},
  volume={36},
  pages={24993--25006},
  year={2023}
}

@inproceedings{shtedritski2023does,
  title={What does clip know about a red circle? visual prompt engineering for vlms},
  author={Shtedritski, Aleksandar and Rupprecht, Christian and Vedaldi, Andrea},
  booktitle={Proceedings of the IEEE/CVF International Conference on Computer Vision},
  pages={11987--11997},
  year={2023}
}

@article{xu2025filmagent,
  title={Filmagent: A multi-agent framework for end-to-end film automation in virtual 3d spaces},
  author={Xu, Zhenran and Wang, Longyue and Wang, Jifang and Li, Zhouyi and Shi, Senbao and Yang, Xue and Wang, Yiyu and Hu, Baotian and Yu, Jun and Zhang, Min},
  journal={arXiv preprint arXiv:2501.12909},
  year={2025}
}

@article{qiu2024moviecharacter,
  title={Moviecharacter: A tuning-free framework for controllable character video synthesis},
  author={Qiu, Di and Chen, Zheng and Wang, Rui and Fan, Mingyuan and Yu, Changqian and Huang, Junshi and Wen, Xiang},
  journal={arXiv preprint arXiv:2410.20974},
  year={2024}
}

@article{bharadhwaj2024gen2act,
  title={Gen2act: Human video generation in novel scenarios enables generalizable robot manipulation},
  author={Bharadhwaj, Homanga and Dwibedi, Debidatta and Gupta, Abhinav and Tulsiani, Shubham and Doersch, Carl and Xiao, Ted and Shah, Dhruv and Xia, Fei and Sadigh, Dorsa and Kirmani, Sean},
  journal={arXiv preprint arXiv:2409.16283},
  year={2024}
}

\newpage
\appendix

\section{Technical Appendix}

\subsection{Baseline Details}
\label{sec:appendix_baselines}

This section provides implementation details for each baseline method used in our experiments. All baseline methods receive the same first-frame image and text prompt as SocialDirector. For methods that require audio input, we supply a 5s silent waveform so that the comparison isolates each model's ability to generate social interaction behaviors from visual and textual cues alone. The model-specific configurations are detailed below.

\begin{itemize}[leftmargin=1.5em]
\item \textbf{Wan2.2}~\citep{wan2025wan} is a state-of-the-art open-source video generation model. We use the I2V-A14B checkpoint (Wan-AI/Wan2.2-I2V-A14B) as our main base model.

\item \textbf{LTX-2.3}~\citep{hacohen2026ltx} is another leading open-source video generation model. We use the 22B checkpoint (Lightricks/LTX-2.3) as a secondary base model to verify generalizability.

\item \textbf{MultiTalk}~\citep{multitalk} is built on Wan2.1-I2V-14B and injects per-speaker audio embeddings into the diffusion backbone. Its inference pipeline natively supports at most two speakers; for scenes with three or more persons, we select the two persons whose action events occur earliest and drop the rest.

\item \textbf{Playmate2}~\citep{playmate2} is built on Wan2.1-I2V-14B-720P and accepts an arbitrary speaker count. Because this method requires separated per-speaker audio inputs, which are unavailable in our multi-person dataset, and its generation speed is prohibitively slow under our setting and hardware, we only include it in qualitative comparisons; for these qualitative cases, we split the native audio track into temporal segments and assign them to 4--6 speakers.

\item \textbf{AnyTalker}~\citep{zhong2025anytalker} is built on the 1.3B Wan2.1-Fun-V1.1-InP backbone and dynamically matches conditioned faces to the input audio list. We feed our annotated per-person bounding boxes together with silent audio tracks into its Audio-Face Attention Mask, so that generation is driven purely by the first frame and text prompt.

\item \textbf{Bind-Your-Avatar}~\citep{huang2025bind} is a CogVideoX-based method that requires exactly two cropped face references by InsightFace. We select the two earliest-event persons in the same way as MultiTalk. Because the model is constrained to 49 frames at 25,fps by CogVideoX's positional embedding, we temporally stretch the output to approximately 5s to match the duration of other baselines.

\item \textbf{HunyuanVideo-Avatar}~\citep{chen2025hunyuanvideo} features a Face-Aware Audio Adapter but its released inference script processes a single image--audio--prompt tuple and only detects the largest face per frame, effectively operating as a single-person baseline on our multi-person scenes.

\item \textbf{EchoMimic~v3}~\citep{meng2026echomimicv3} is a single-subject audio-driven portrait animation method based on Wan2.1-Fun-1.3B, included as a high-quality single-person baseline.
\end{itemize}

\subsection{Evaluation Details}
\label{sec:appendix_eval}

This section provides additional details on the evaluation pipeline introduced in \cref{sec:eval}, including the generation prompt format, action category taxonomy, and VLM query design.

\paragraph{Generation Prompt Format.}
To generate videos for evaluation, we construct a structured text prompt for each sample. The prompt begins with a scene-level description enumerating all speakers by their left-to-right position, followed by one sentence per speaker describing their action. Below are two representative examples with \textit{directional words} emphasized:

\begin{itemize}[leftmargin=1.5em]
\item ``There are 3 people in the scene: the person on the left (speaker 1), the person in the middle (speaker 2), the person on the right (speaker 3). [0s--3s] The person on the left listening while touching his chin. [1s--4s] The person in the middle \textit{speaks leftward} to speaker 1 with anger. The person on the right remains still with no notable action.''
\item ``There are 2 people in the scene: the person on the left (speaker 1), the person on the right (speaker 2). [0s--4s] The person on the left speaking while waving his hand. [2s--5s] The person on the right smiling with joy.''
\end{itemize}

\paragraph{Action Categories.}
Each annotated event is associated with a free-form action description in our datasets. To enable systematic evaluation for action accuracy, we map these descriptions to one of 11 coarse action categories via keyword matching. Each category corresponds to a fixed yes/no question template used by the VLM evaluator, as listed in \cref{tab:action_categories}.

\begin{table}[h]
\centering
\small
\caption{Action categories and their VLM question templates. Each action accuracy query follows the format: ``In this video, does \{positional description\} \{template\}? Answer Yes or No.''}
\label{tab:action_categories}
\setlength{\tabcolsep}{6pt}
\begin{tabular}{cll}
\Xhline{1.0pt}
\rowcolor[gray]{0.92}
\bf \# & \bf Category & \bf Question template fragment \\
\hline
1  & pointing            & ``point at someone or something'' \\
2  & object interaction  & ``interact with an object (e.g.\ pick up, put down, hold)'' \\
3  & head gesture        & ``make a head gesture (e.g.\ nod or shake head)'' \\
4  & mutual gesture      & ``physically interact with another person or clap'' \\
5  & body posture        & ``change body posture (e.g.\ cross arms, lean, stand up)'' \\
6  & speaking            & ``appear to be speaking (e.g.\ mouth moving)'' \\
7  & facial expression   & ``show a facial expression (e.g.\ smile or laugh)'' \\
8  & listening           & ``listen attentively'' \\
9  & looking             & ``look at someone or something'' \\
10 & hand gesture        & ``make a hand gesture (e.g.\ wave, raise hand)'' \\
11 & drinking/toasting   & ``drink or make a toast'' \\
\Xhline{1.0pt}
\end{tabular}
\end{table}

\paragraph{VLM Query Design.}
As introduced in \cref{sec:eval_metrics}, we formulate each metric as a binary VQA task. Each query is evaluated on a temporally cropped sub-clip $[\text{event.start} - 1\text{s},\; \text{event.end} + 1\text{s}]$. Speakers are identified via colored bounding boxes drawn on the video frames, and questions refer to ``the person in the red/green box'' rather than positional descriptions. Representative queries for each metric are listed below:

\begin{itemize}[leftmargin=1.5em]
\item \textbf{Action accuracy}: ``In this video, does the person in the red box interact with any object? Answer Yes or No.'' (Expected: Yes)
\item \textbf{Target accuracy}: ``In this video, does the person in the red box point toward the person in the green box? Answer Yes or No.'' (Expected: Yes)
\item \textbf{Stillness accuracy}: ``In this video, does the person in the red box perform any notable gesture or directed action? Answer Yes or No.'' (Expected: No)
\end{itemize}

\subsection{Reliability Analysis of Social Interaction Metrics}
\label{sec:appendix_reliability}
 
To verify that the improvements of SocialDirector reported in \cref{sec:results} are robust under our VLM-based evaluation pipeline rather than artifacts of a particular evaluator or random seed, we provide two complementary analyses: (i) per-VLM scores prior to majority voting, and (ii) per-seed scores across three random seeds.
 
\paragraph{Per-VLM Consistency.}
\cref{tab:appendix_per_vlm} reports the social interaction metrics evaluated by each of the three VLMs (Qwen3-VL, InternVL3.5, MiniCPM-V) independently, before majority voting. Across all three evaluators, SocialDirector consistently improves over the Wan2.2 base model on all three social interaction metrics, indicating that the gains are not attributable to the bias of any single VLM. 
 
\begin{table}[h]
\centering
\small
\caption{Per-VLM social interaction metrics (\%) without majority voting. \textcolor{bettergreen}{Green} indicates improvement over the base model.}
\label{tab:appendix_per_vlm}
\setlength{\tabcolsep}{4pt}
\resizebox{\linewidth}{!}{
\begin{tabular}{l|ccc|ccc|ccc}
\Xhline{1.0pt}
\rowcolor[gray]{0.92}
 & \multicolumn{3}{c|}{\bf Qwen3-VL} & \multicolumn{3}{c|}{\bf InternVL3.5} & \multicolumn{3}{c}{\bf MiniCPM-V 4.5} \\
\rowcolor[gray]{0.92}
\multirow{-2}{*}{\bf Method} & Action $\uparrow$ & Target $\uparrow$ & Stillness $\uparrow$ & Action $\uparrow$ & Target $\uparrow$ & Stillness $\uparrow$ & Action $\uparrow$ & Target $\uparrow$ & Stillness $\uparrow$ \\
\hline
\color{gray} GT Oracle & \color{gray} 72.9 & \color{gray} 75.9 & \color{gray} 77.4 & \color{gray} 74.9 & \color{gray} 63.1 & \color{gray} 97.8 & \color{gray} 72.7 & \color{gray} 69.7 & \color{gray} 83.1 \\
\hline
Wan2.2 & 69.8 & 73.9 & 73.5 & 72.8 & 61.7 & 95.0 & 70.2 & 67.6 & 74.6 \\
\textbf{Wan2.2 + SocialDirector} & 73.5\makebox[0pt][l]{\scriptsize\textcolor{bettergreen}{\,+3.7}} & 77.3\makebox[0pt][l]{\scriptsize\textcolor{bettergreen}{\,+3.4}} & 77.4\makebox[0pt][l]{\scriptsize\textcolor{bettergreen}{\,+3.9}} & 75.7\makebox[0pt][l]{\scriptsize\textcolor{bettergreen}{\,+2.9}} & 65.6\makebox[0pt][l]{\scriptsize\textcolor{bettergreen}{\,+3.9}} & 95.5\makebox[0pt][l]{\scriptsize\textcolor{bettergreen}{\,+0.5}} & 73.9\makebox[0pt][l]{\scriptsize\textcolor{bettergreen}{\,+3.7}} & 69.7\makebox[0pt][l]{\scriptsize\textcolor{bettergreen}{\,+2.1}} & 79.6\makebox[0pt][l]{\scriptsize\textcolor{bettergreen}{\,+5.0}} \\
\Xhline{1.0pt}
\end{tabular}
}
\end{table}
 
\paragraph{Per-Seed Robustness.}
\cref{tab:appendix_per_seed} reports the mean and standard deviation of social interaction metrics across three random seeds. On both base models, SocialDirector's gains on action and target accuracy clearly exceed the seed-level standard deviations, confirming that the improvements are statistically robust rather than seed-specific artifacts.
 
\begin{table}[h]
\centering
\small
\caption{Per-seed social interaction metrics (\%, mean $\pm$ std over 3 seeds).}
\label{tab:appendix_per_seed}
\setlength{\tabcolsep}{6pt}
\begin{tabular}{l|ccc}
\Xhline{1.0pt}
\rowcolor[gray]{0.92}
\bf Method & Action Acc $\uparrow$ & Target Acc $\uparrow$ & Stillness Acc $\uparrow$ \\
\hline
LTX-2.3                       & 71.6 $\pm$ 1.3 & 69.4 $\pm$ 0.5 & 87.6 $\pm$ 0.9 \\
\textbf{LTX-2.3 + SocialDirector} & 74.3 $\pm$ 1.3 & 70.8 $\pm$ 0.5 & 88.3 $\pm$ 1.5 \\
\Xhline{0.5pt}
Wan2.2                       & 72.2 $\pm$ 1.5 & 69.2 $\pm$ 1.6 & 85.1 $\pm$ 1.3 \\
\textbf{Wan2.2 + SocialDirector} & 76.1 $\pm$ 0.7 & 72.3 $\pm$ 0.8 & 88.2 $\pm$ 2.3 \\
\Xhline{1.0pt}
\end{tabular}
\end{table}

\subsection{More Visualization Results}
\label{sec:appendix_more_vis}

\begin{figure*}[h]
\centering
\includegraphics[width=\linewidth]{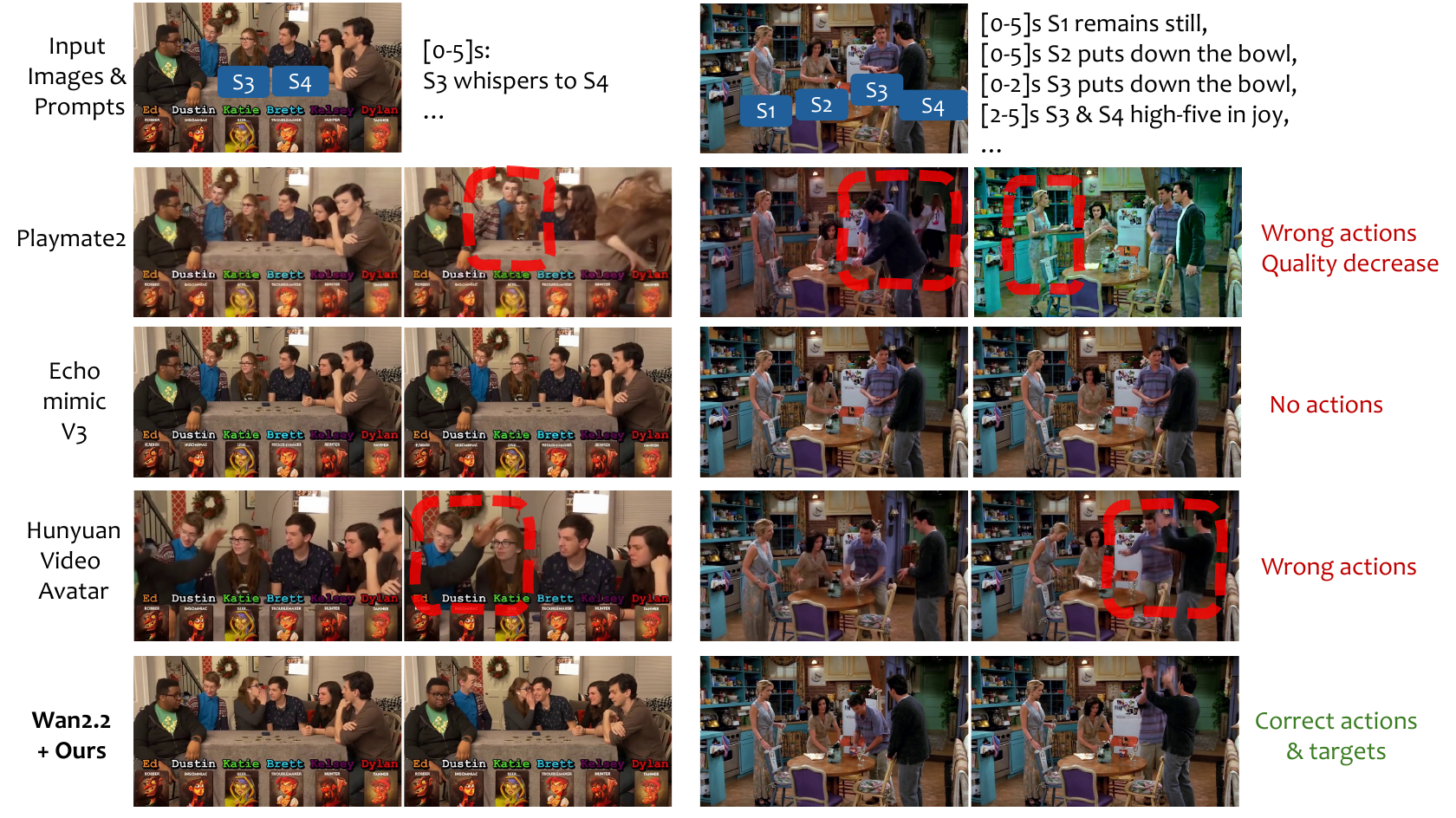}
\caption{Additional qualitative comparisons of SocialDirector against baseline methods.}
\label{fig:more_vis}
\end{figure*}

\cref{fig:more_vis} presents additional qualitative comparisons of SocialDirector against a broader set of baseline methods on diverse social interaction scenarios. Playmate2~\citep{playmate2} suffers from severe video quality degradation when applied to multi-person scenes with non-separated audio. EchoMimic~v3~\citep{meng2026echomimicv3} produces nearly static videos with little interaction-relevant motion. HunyuanVideo-Avatar~\citep{chen2025hunyuanvideo} likewise fails to handle multi-person scenes: in the left example it zooms in on S3 and S4, while in the right example it exhibits incorrect action--actor binding. Under the same inputs, SocialDirector correctly generates the intended interactions with accurate actor--action--target assignment and consistent identity across speakers. 


\subsection{Broader Impacts}
\label{sec:appendix_impact}
 
SocialDirector advances controllable multi-person video generation with potential positive impacts in film production, social robotics, virtual avatars, and accessibility applications such as creating engaging educational and communication content. As a human-centric video generation method, however, it also shares the broader risks of generative video technologies, including the possibility of being misused to synthesize misleading or ethically inappropriate content involving multiple individuals. Our research is conducted entirely on publicly available datasets and open-source pretrained models with no additional video data collection. To mitigate the risk of misuse, we preserve the built-in watermarks of the underlying base models (Wan2.2 and LTX-2.3) in all generated videos, so that outputs of our pipeline remain identifiable as synthetic content.

\end{document}